\documentclass[sigconf,timestamp,screen]{acmart}
\renewcommand\footnotetextcopyrightpermission[1]{} 

\setcitestyle{square}

\usepackage{arydshln}
\usepackage{soul}
\usepackage{mathtools}
\usepackage{siunitx}
\usepackage{multirow}
\usepackage{hyperref}
\usepackage{footmisc}
\usepackage[ruled]{algorithm2e} 
\usepackage{booktabs} 
\usepackage{amsmath}
\usepackage{graphicx}
\usepackage{subcaption}
\usepackage{amsmath}

\SetAlFnt{\small}
\SetAlCapFnt{\small}
\SetAlCapNameFnt{\small}
\SetAlCapHSkip{0pt}
\IncMargin{-\parindent}

\acmDOI{}

\received{}

\begin{document}

\title{The joint role of geometry and illumination on material recognition}

\author{Manuel Lagunas}
\authornote{Corresponding author, mlagunas@unizar.es}
\affiliation{%
  \institution{Universidad de Zaragoza, I3A}
  \city{Zaragoza, Spain}
}

\author{Ana Serrano}
\affiliation{%
  \institution{Universidad de Zaragoza, I3A, Max Planck Institute for Informatics}
  \city{Zaragoza, Spain}
}

\author{Diego Gutierrez}
\affiliation{%
  \institution{Universidad de Zaragoza, I3A}
  \city{Zaragoza, Spain}
}

\author{Belen Masia}
\affiliation{%
  \institution{Universidad de Zaragoza, I3A}
  \country{Zaragoza, Spain}
}

\begin{abstract}
Observing and recognizing materials is a fundamental part of our daily life. Under typical viewing conditions, we are capable of effortlessly identifying the objects that surround us and recognizing the materials they are made of. Nevertheless, understanding the underlying perceptual processes that take place to accurately discern the visual properties of an object is a long-standing problem. In this work, we perform a comprehensive and systematic analysis of how the interplay of geometry, illumination, and their spatial frequencies affects human performance on material recognition tasks. We carry out large-scale behavioral experiments where participants are asked to recognize different reference materials among a pool of candidate samples. In the different experiments, we carefully sample the information in the frequency domain of the stimuli. From our analysis, we find significant first-order interactions between the geometry and the illumination, of both the reference and the candidates. In addition, we observe that simple image statistics and higher-order image histograms do not correlate with human performance. Therefore, we perform a high-level comparison of highly non-linear statistics by training a deep neural network on material recognition tasks. Our results show that such models can accurately classify materials, which suggests that they are capable of defining a meaningful representation of material appearance from labeled proximal image data. Last, we find preliminary evidence that these highly non-linear models and humans may use similar high-level factors for material recognition tasks.
\end{abstract}


\maketitle

\setlength\dashlinedash{0.2pt}
\setlength\dashlinegap{2pt}
\setlength\arrayrulewidth{0.2pt}
\begin{sloppypar}

\section{Introduction}

Under typical viewing conditions, humans are capable of effortlessly recognizing materials and inferring many of their key physical properties, just by briefly looking at them. While this is almost an effortless process, it is not a trivial task. The image that is input to our visual system results from a complex combination of the surface geometry, the reflectance of the material, the distribution of lights in the environment, and the observer's point of view. 
To recognize the material of a surface while being invariant to other factors of the scene, our visual system carries out an underlying perceptual process that is not yet fully understood~\cite{dror2001estimating,fleming2001humans,adelson2000lightness}.

So how does our brain recognize materials?
We could think that, similar to solving an inverse optics problem, our brain is estimating the physical properties of each material~\cite{pizlo2001perception}. This would imply knowledge of many other physical quantities about the object and its surrounding scene, from which our brain could disentangle the reflectance of the surface. 
However, we rarely have access to such precise information, so variations based on Bayesian inference have been proposed~\cite{kersten2004object}.

Other approaches are based on image statistics, and explain material recognition as a process where our brain extracts image features that are relevant to describe materials. Then, it would try to match them with previously acquired knowledge, in order to discern the material we are observing. Considering this approach our visual system would disregard the illumination, motion, or other factors in the scene and try to recognize materials by representing their typical appearance in terms of features instead of explicitly acquiring an accurate physical description of each factor. This type of image analysis can be carried out in the primary domain~\cite{geisler2008visual,adelson2008image,fleming2014,motoyoshi2007image,nishida1998use}, or in the frequency domain~\cite{oliva2001modeling,brady2012spatial,giesel2013frequency}. However, it is argued if our visual system actually derives any aspects of material perception from such simple statistics~\cite{anderson2009image}. For instance, Fleming and Storrs~\shortcite{fleming2019learning} have recently proposed the idea that highly non-linear encodings of the visual input may better explain the underlying processes of material perception.

\begin{figure}[t]
	\centering
	\includegraphics[width=\linewidth]{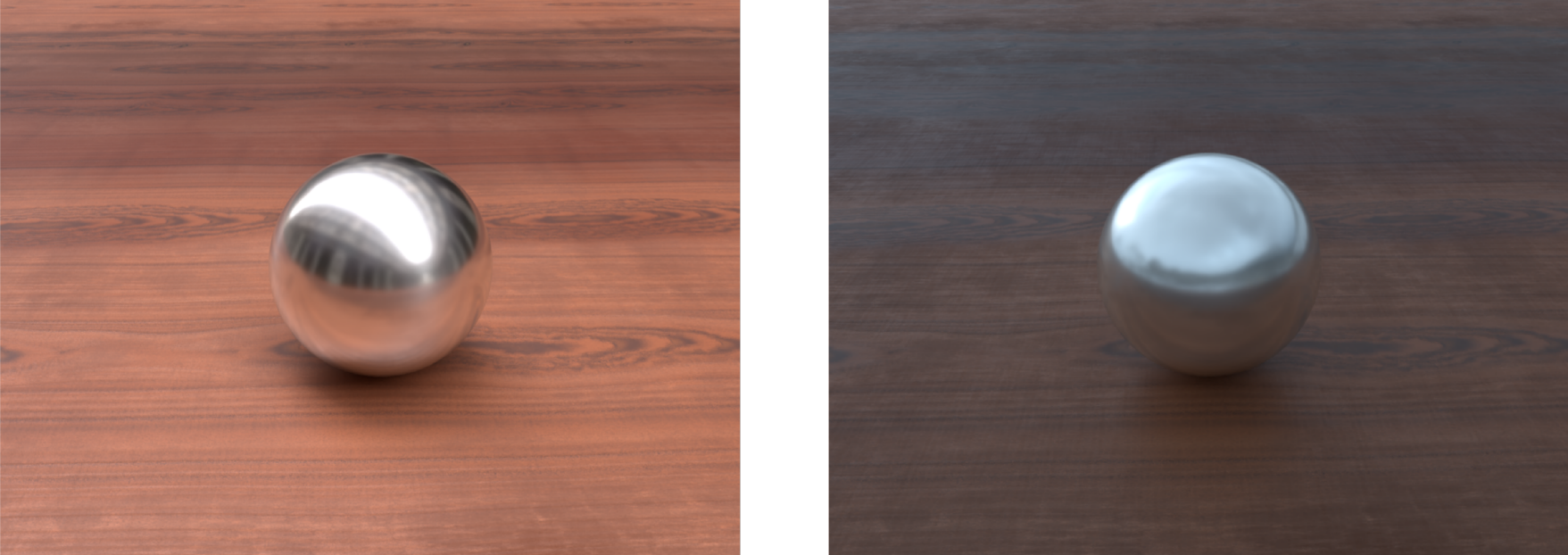}
	\caption{Two spheres made of silver, under two different illuminations, leading to completely different pixel-level statistics.}
	\label{fig:img_light_comparison}
\end{figure}

\begin{figure}[t]
	\centering
	\includegraphics[width=\linewidth]{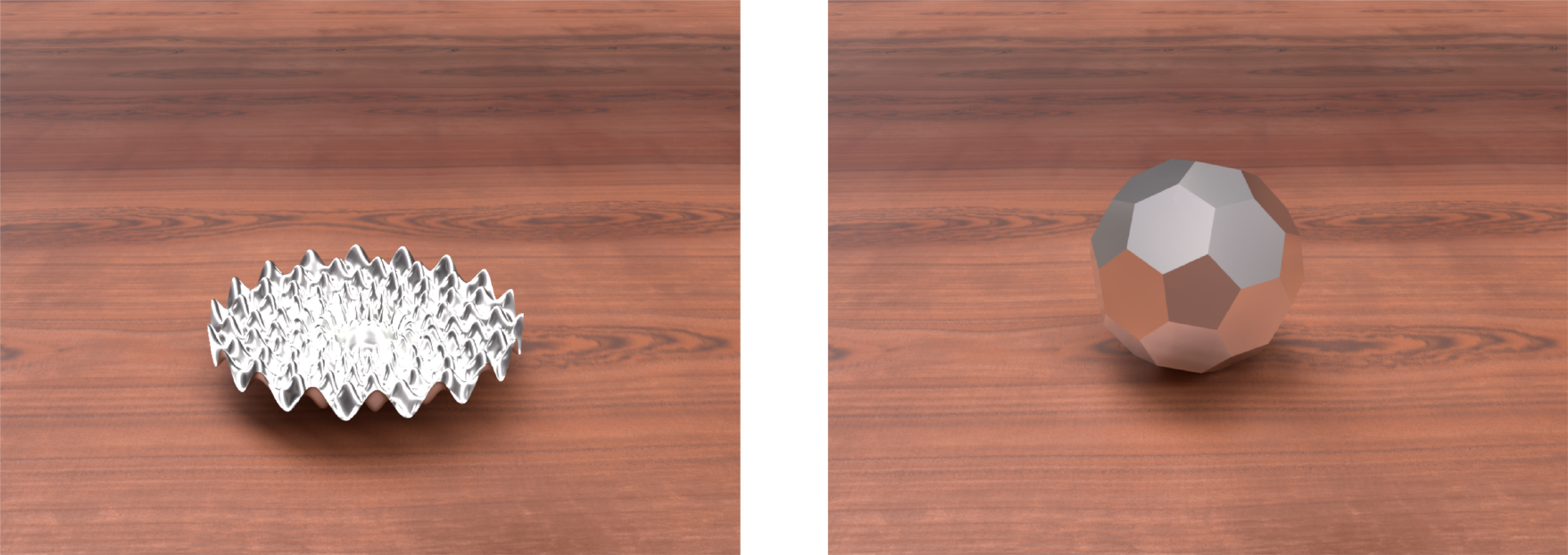}
	\caption{Two objects of different geometries but made of the same material, under the same illumination. The object on the left appears to be made of a shinier material.}
	\label{fig:img_geom_comparison}
\end{figure}

In this work, we thoroughly analyze how the confounding effects of illumination and geometry influence human performance in material recognition tasks. The same material can yield different appearances due to changes in illumination and/or geometry (see Figures~\ref{fig:img_light_comparison} and~\ref{fig:img_geom_comparison}), while it is possible to have two different materials look the same by tweaking the two parameters~\cite{vangorp2007}. 
We aim to further our understanding of the complex interplay between geometry and illumination in material recognition. 
We have carried out large-scale, rigorous online behavioral experiments where participants were asked to recognize different materials, given images of one reference material and a pool of candidates. By using photorealistic computer graphics, we obtain carefully controlled stimuli, with varying degrees of information in the frequency domain.
In addition, we observe that simple image statistics, image histograms, and histograms of V1-like subband filters do not correlate with human performance in material recognition tasks. Inspired by Fleming and Storrs' recent work~\shortcite{fleming2019learning}, we analyze highly non-linear statistics by training a deep neural network. We observe that such statistics define a robust and accurate representation of material appearance and find preliminary evidence that these models and humans may share similar high-level factors when recognizing materials.

\subsection{Material recognition}

Recognizing materials and inferring their key features by sight is invaluable for many tasks. Our experience suggests that humans are able to correctly predict a wide variety of rough material categories like textiles, stones, or metals~\cite{fleming2014, ged2010recognizing,fleming2005low,li2012recognizing}; or items that we would call "stuff"~\cite{adelson2001seeing} --- like sand or snow---. Humans are also capable of identifying the materials in a photograph by briefly looking at them~\cite{sharan2009, sharan2008eye} or of inferring their physical properties without the need to touch them~\cite{fleming2013perceptual, fleming2015short,maloney2010, nagai2015,serrano2016,jarabo2014btf}. This ability is built from experience, by actually confirming visual impressions with other senses. This way, material perception becomes a cognitive process~\cite{palmer1975} whose underlying intricacies are not fully understood yet~\cite{anderson2011, fleming2015, Thompson2011}.

\subsection{Interplay of geometry and illumination}
Material perception is a complex process that involves a large number of distinct dimensions~\cite{seve1993problems, obein2004difference, mao2019motion} that, sometimes, are impossible to physically measure~\cite{hunter1937methods}. The illumination of a scene~\cite{zhang2015influence, beck1981highlights, bousseau2011optimizing} and the shape of a surface, are responsible for the final appearance of an object~\cite{vangorp2007, schluter2019visual,nishida1998use}, and therefore, for our perception of the materials it is made of~\cite{olkkonen2011joint}. 
Humans are capable of estimating the reflectance properties of a surface~\cite{dror2001surface} even when there is no information about its illumination~\cite{dror2001estimating, fleming2001humans}, yet we perform better under illuminations that match real-world statistics~\cite{fleming2003real}. 
Indeed, geometry and illumination have a joint interaction in our perception of glossiness~\cite{marlow2012perception, olkkonen2011joint,leloup2010geometry, faul2019influence}, and color~\cite{bloj1999perception}. In this work, we explore the interplay of shape, illumination, and their spatial frequencies in our performance at recognizing materials. To achieve that, we launch rigorous online behavioral experiments where we rely on realistic computer graphics to generate the stimuli and carefully vary their information in the frequency domain. 

\subsection{Image statistics and material perception}
One of the goals in material perception research is to untangle the processes that happen on our visual system in order to comprehend their roles and know what information they carry. There is an ongoing discussion on whether our visual system is solving an inverse optics problem~\cite{kawato1993forward, pizlo2001perception} or if it matches the statistics of the input to our visual system~\cite{adelson2000lightness,motoyoshi2007image,thompson2016visual} in order to understand the world that surrounds us.
Later studies regarding our visual system and how we perceive materials dismiss the inverse optics approach and claim that it is unlikely that our brain estimates the parameters of the reflectance of a surface, when, for instance, we want to measure glossiness~\cite{fleming2014, geisler2008visual}. Instead, they suggest that our visual system joins low and mid-level statistics to make judgments about surface properties~\cite{adelson2008image}. On this hypothesis, Motoyoshi et al.~\shortcite{motoyoshi2007image} suggest that the human visual system could be using some sort of measure of histogram symmetry to distinguish glossy surfaces. 
Other works have explored image statistics in the frequency domain~\cite{hawken1987spatial,schiller1976quantitative}, for instance, to characterize material properties~\cite{giesel2013frequency}, or to discriminate textures~\cite{julesz1962visual, schaffalitzky2001viewpoint}. 
However, it is argued if our visual system actually derives any aspects of material perception from simple statistics~\cite{anderson2009image, kim2010image, olkkonen2010perceived}. Instead, recent work by Fleming and Storrs~\shortcite{fleming2019learning} proposes that to infer the properties of the scene, our visual system is doing an efficient and accurate encoding of the proximal stimulus (image input to our visual system). Thus, highly non-linear models, such as deep neural networks, may better explain human perception. 
In line with such observation, Bell et al.~\shortcite{bell2015material} show how deep neural networks can be trained in a supervised fashion to accurately recognize materials, and Wang et al.~\shortcite{wang20164d} later extend it to also recognize materials in light fields. 
Closer to ours, Lagunas et al.~\shortcite{lagunas2019similarity} devise a deep learning based material similarity metric that correlates with human perception. They collected judgements on perceived material similarity as a whole, not explicitly taking into account the influence of geometry or illumination, and build their metric upon such judgements. 
In contrast, we focus on analyzing to which extent  geometry and illumination do interfere with our perception of material appearance. We launch several behavioral experiments with carefully controlled stimuli, and ask participants to specify which materials are closer to a reference. 
In addition, taking inspiration from these recent works, we explore how highly non-linear models, such as deep neural networks, perform in material classification tasks. 
We find that such models are capable of accurately recognizing materials, and further observe that deep neural networks may share similar high-level factors to humans when recognizing materials.

\section{Methods}
\label{sec:user study}

\begin{figure}[t]
	\centering
	\includegraphics[width=\columnwidth]{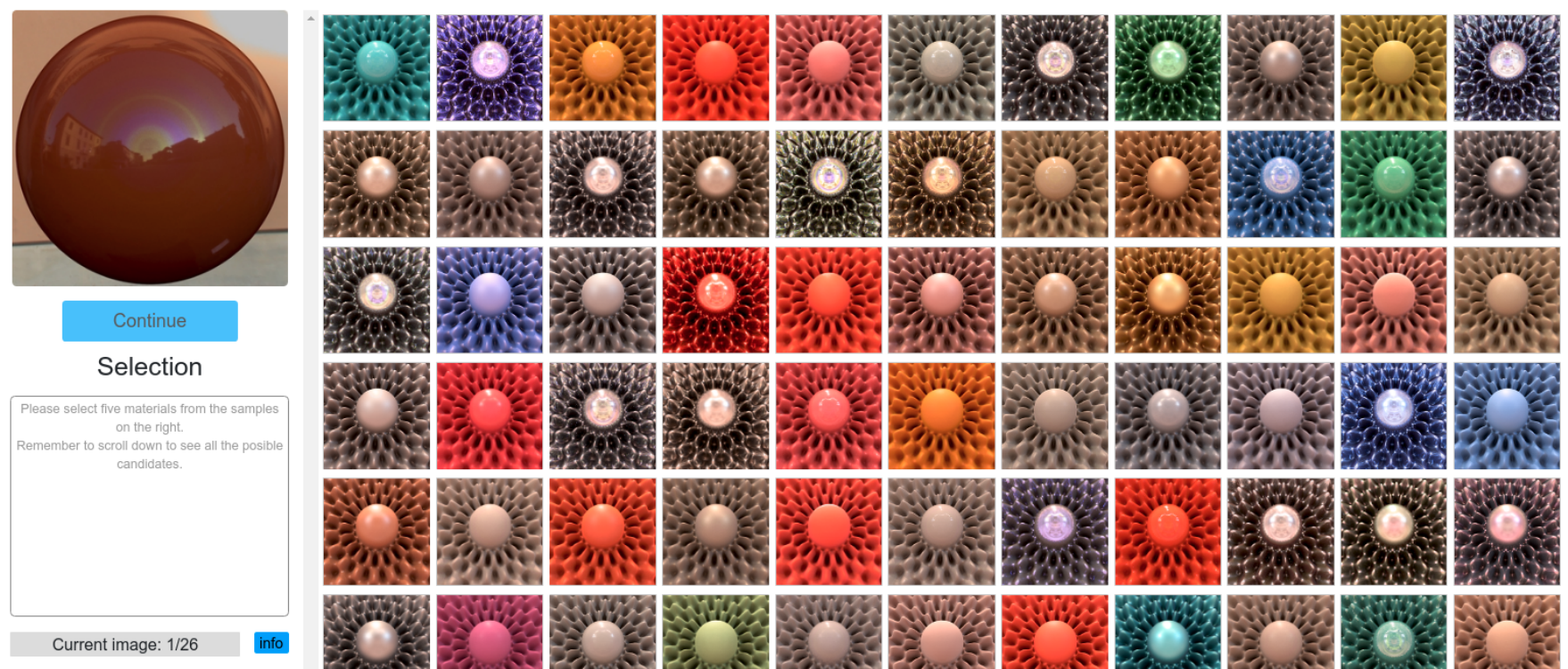}
	\caption{Graphical user interface of the online behavioral experiments. In particular, this screenshot belongs to the \textsc{Test SH}. On the left, the user can see the reference material together with her current selection. On the right, she can observe all the candidate materials. To select one candidate material, the user clicks on the corresponding image and it is automatically added to the selection box on the left.}
	\label{fig:test_screenshot}
\end{figure}

We carry out a set of online behavioral experiments where we analyze the influence of geometry, illumination, and their frequencies in human performance for material recognition tasks.
Participants are presented with a reference material and their main task is to pick five materials from a pool of candidates that they think are closer to the reference. A screenshot of the experiment can be seen in Figure~\ref{fig:test_screenshot}.

\subsection{Stimuli}
We obtain our stimuli from the dataset proposed by Lagunas et al.~\shortcite{lagunas2019similarity}. This dataset contains images created using photorealistic computer graphics, with  15 different geometries; six different real-world illuminations ranging from indoor scenarios to urban or natural landscapes; and 100 different materials measured from their real-world counterparts which were pooled from MERL database~\cite{matusik2003}.
We sample the following factors for our experiments:

\paragraph{Geometries}
Among the geometries that the dataset contains, we choose the sphere and {\emph{Havran-2}} geometry~\cite{havran2016}. These are a low and high spatial frequency geometries, respectively, suitable to test how the spatial frequencies of the geometry affect the final appearance of the material and our performance at recognizing it.
\begin{itemize}
	\item \textit{Sphere:} Representing a smooth, and low spatial frequency geometry, widely adopted in previous behavioral experiments~\cite{filip2008psychophysically, jarabo2014btf, kerr2010toward,sun2017attribute}.
	\item \textit{Havran-2:}\footnote{To simplify the notation we will refer to this geometry as \emph{Havran}.} It is a geometry with high spatial frequencies, and with high spatial variations that has been obtained through optimization techniques.
	\emph{Havran-2} surface has had significant success in recent perceptual studies and applications~\cite{lagunas2019similarity, guo2018brdf, vavra2016minimal, guarnera2018perceptually}.
\end{itemize}

\begin{figure}[t]
	\centering 
	\includegraphics[width=\columnwidth]{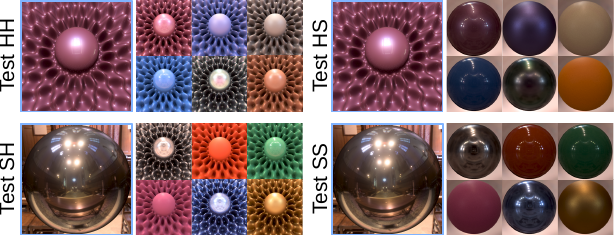}
	\caption{Examples of the stimuli in each different online behavioral experiment. On the left, we show an example of the reference stimuli with one of the six illuminations. On the right, we show a small subset (six out of the 100 materials) of the candidate stimuli with \emph{St. Peters} illumination.}
	\label{fig:test-examples}
\end{figure}

The stimuli in each different experiment can be observed in Figure~\ref{fig:test-examples}. The geometry in the reference and candidate samples changes depending on the experiment, the details are as follows:
\begin{itemize}
	\item \textsc{Test HH:} Both the reference and the candidates depict \emph{Havran} geometry.
	\item \textsc{Test HS:} The reference depicts \emph{Havran} and the candidates depict the sphere.
	\item \textsc{Test SH:} The reference depicts the sphere while the candidates depict \emph{Havran}.
	\item \textsc{Test SS:} Both the reference and the candidates depict the sphere geometry.
\end{itemize}

\begin{figure}[t]
	\centering
	\includegraphics[width=\columnwidth]{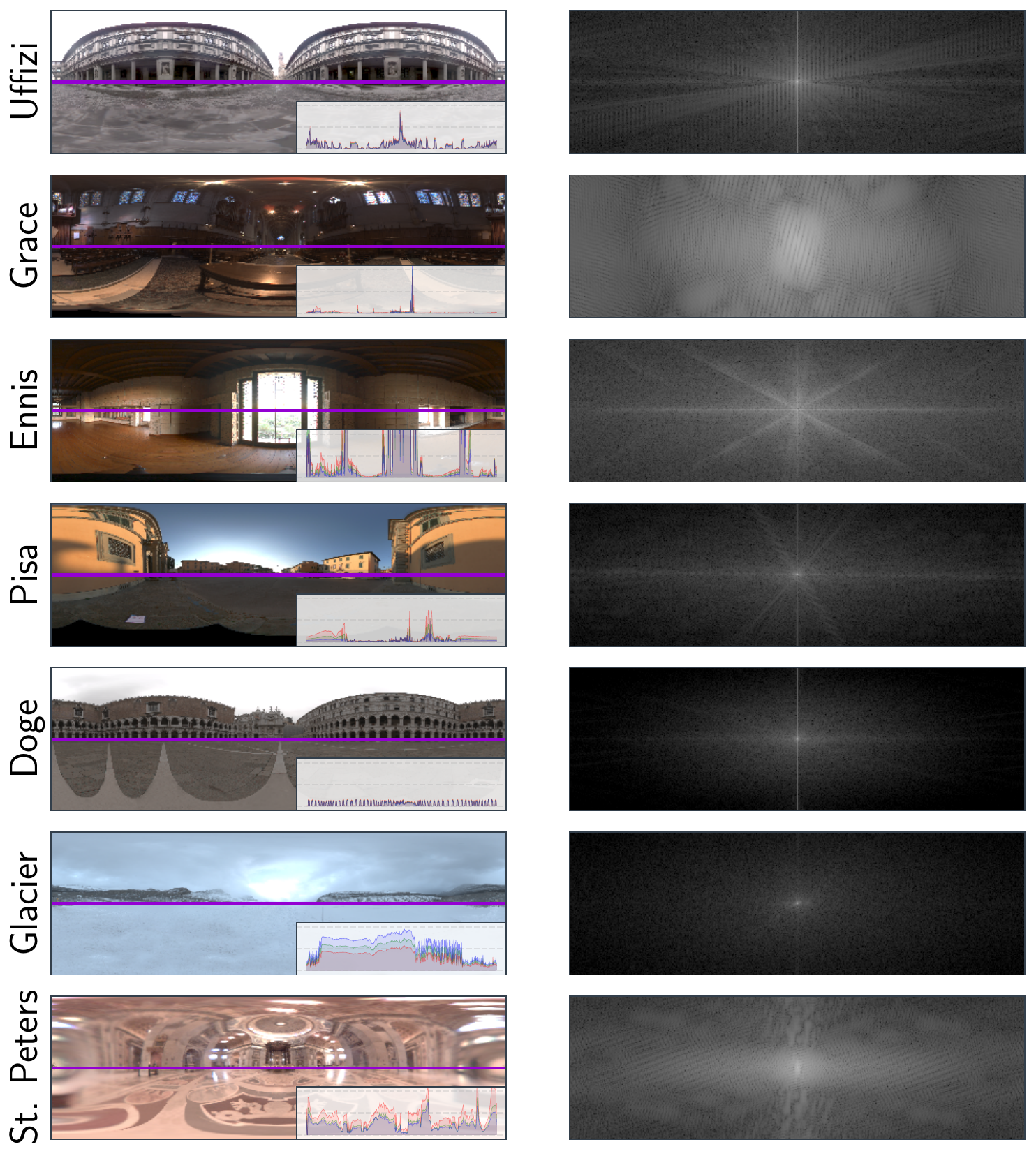}
	\caption{\emph{Left:} All illuminations depicted in the online behavioral experiments. The inset corresponds to the pixel intensity for the horizontal purple line. \emph{Right:} Magnitude spectrum of the luminance for each illumination.}
	\label{fig:ref_illum}
\end{figure}

\begin{figure}[t]
	\centering
	\includegraphics[width=0.9\columnwidth]{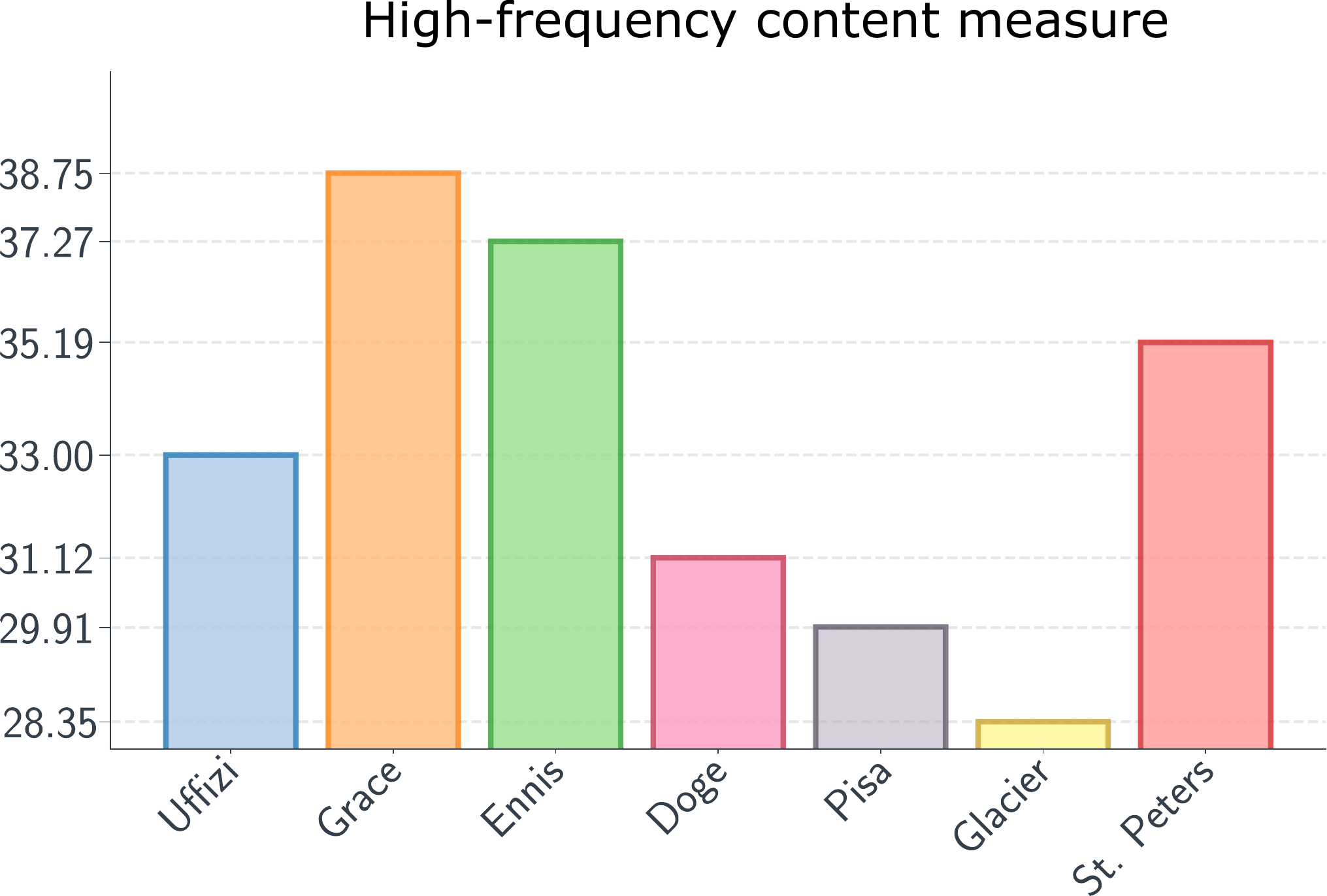}
	\caption{High-frequency content (HFC) measure computed for all the candidate and reference illuminations. We can observe how high-frequency illuminations (\emph{Uffizi}, \emph{Grace}, \emph{Ennis}, \emph{St. Peters}) also have a high HFC value, medium-frequency illuminations (\emph{Pisa}, \emph{Doge}) have a lower HFC value, and, last, low-frequency illuminations (\emph{Glacier}) have the lowest HFC value.}
	\label{fig:hfc}
\end{figure}

\paragraph{Illuminations}
To prevent a pure matching task, we choose different illuminations between the reference and candidate materials for all behavioral experiments.
\begin{itemize}
	\item
		  The reference samples depict six different illuminations captured from the real-world. All illuminations can be observed in Figure~\ref{fig:ref_illum}. To have an intuition of the content in the captured illumination, the insets show the RGB intensity for the horizontal purple line. We use all illuminations in the dataset since they contain a mix of spatial frequencies suitable to empirically test how the spatial frequencies of the illumination may affect human performance on material recognition tasks. 
	      The illuminations \emph{Grace}, \emph{Ennis}, and \emph{Uffizi} have a broad spatial frequency spectrum, \emph{Pisa} and \emph{Doge} mainly contain medium and low spatial frequency content, while \emph{Glacier} mainly has low spatial frequency content. To simplify the notation, we will refer to them throughout the paper as high-frequency, medium-frequency, and low-frequency illuminations, respectively.
	\item
		  The candidate samples depict the \textit{St. Peters} illumination (except in an additional experiment discussed in Section~\ref{sec:discussion} where they depict \emph{Doge} illumination). \emph{St. Peters} is an illumination  that has been used in the past for several perceptual studies~\cite{fleming2003real,serrano2016}, and it can be seen in Figure~\ref{fig:ref_illum}. The inset shows the RGB pixel intensity for the horizontal purple line. 
\end{itemize}
To quantify the spatial frequencies of the illuminations, we have employed the high-frequency content measure (HFC)~\cite{brossier2004real}. This measure characterizes the frequencies in a signal by summing linearly-weighted values of the spectral magnitude, thus avoiding to arbitrarily choose a separation between high and low frequencies, or visually assessing the slope of the 1/f amplitude spectrum. A high HFC value means higher frequencies in the signal. Figure~\ref{fig:hfc} shows the HFC for each illumination.

\paragraph{Materials} We use all the materials from the Lagunas et al. dataset~\shortcite{lagunas2019similarity}. The reference trials are sampled uniformly to cover all 100 material samples in the dataset.
Examples of the stimuli used in each behavioral experiment are shown in Figure~\ref{fig:test-examples}, where the image on the left shows the reference material and the right area shows a subset of the candidate materials.

\subsection{Participants}
The online behavioral experiments were designed to work across platforms on standard web browsers, and they were conducted through the Amazon Mechanical Turk (MTurk) platform. In total, 847 unique users took part in them (368 users belonging to the experiments explained in Section~\ref{sec:analysis}, and 479 belonging to the additional experiments explained in Section~\ref{sec:discussion}), 44.61\% of them female. Among the participants, 62.47\% claimed to be familiar with computer graphics, 25.57\% had no previous experience and 9.96\% declared themselves professionals. 
We also sampled data regarding the devices used during the experiments: 94.10\%  used a monitor, 4.30\% used a tablet, and 1.60\% used a mobile phone. In addition, the most common screen size was 1366x728 px. (42.01\% of participants), minimum screen size was 640x360 px. (two people), and a maximum of 2560x1414 px. (one person).
Users were not aware of the purpose of the behavioral experiment.  

\subsection{Procedure}
Subjects are shown a reference sample and a group of candidate material samples. Each experiment, \emph{HIT} in MTurk terminology, consists of 23 unique reference material samples or \emph{trials}, three of which are sentinels used to detect malicious or lazy users. Users are asked to "\emph{select five material samples which you believe are closer to the one shown in the reference image}". Additionally, we instruct them to make their selection in decreasing order of confidence. We let the users pick five candidate materials because just one answer would provide sparse results. We launched 25 \emph{HITs} for each experiment and each \emph{HIT} was answered by six different users. This resulted in a total of 27.000 non-sentinel trials, 12.000 belonging to the four experiments analyzed in Section~\ref{sec:analysis}, and 15.000 of them belonging to the five additional experiments discussed in Section~\ref{sec:discussion} (a total of nine different experiments with 25 \emph{HITs} each, each \emph{HIT} answered by six users and 20 non-sentinel trials per \emph{HIT}). Users were not allowed to repeat the same \emph{HIT}.

The set of materials in the candidate samples does not vary across \emph{HITs}, however, the position of each sample is randomized for each trial. This has a two-fold purpose: it prevents the user from memorizing the position of the samples, and it prevents them from selecting only the candidate samples that appear at the top of their screen. 
The reference samples do not repeat materials during a \emph{HIT} and the reference material is always present among the candidate samples. During the experiment, stimuli keep a constant display size of 300x300 px. for the reference, and of 120x120 px. for the candidate stimuli (except for some of the additional experiments explained in Section~\ref{sec:discussion} where both reference and candidate stimuli are displayed at either 300x300 px. or 120x120 px.).
Figure~\ref{fig:test_screenshot} shows a screenshot with the graphical user interface during the behavioral experiments. On the left-hand side, we can observe the selection panel with the current trial and the selection of the current materials. The right-hand side displays the set of candidate materials whereof users can pick their selection. Users were not able to go back and re-do an already answered trial, but they could edit their current selection of five materials until they were satisfied with their choice.  Additionally, once the 23 trials of the \emph{HIT} are answered, to have an intuition about the main features that humans use for material recognition, we asked the user: \textit{"Which visual cues did you consider to perform the test?"}.

To minimize worker unreliability, the user performs a brief training before the real test~\cite{welinder2010multidimensional}. To avoid giving the user further information about the test, we use a different geometry (\emph{Havran-3}~\cite{havran2016}) during the training phase.
In this phase, the items of the interface are explained and the user is given guidance on how to perform the test using just a few images~\cite{GarcesSIG2014,Rubinstein10Comparative,lagunas2018}.

\paragraph{Sentinels} Each sentinel shows a randomly selected image from the pool of candidates as the reference sample. We consider user answers to the sentinel as valid if they pick the right material within their five selections, regardless of the order.
We rejected users that did not correctly answer at least one out of the three sentinel questions. 
In order to ensure that users answers were well thought and that they were paying attention to the experiment, we also rejected users that took less than five seconds per trial (on average). 
In the end, we adopt a conservative approach and rejected 19.8\% of the participants, gathering 21.660 answers (9.560 belonging to the behavioral experiments explained in Section~\ref{sec:analysis} and 12.100 belonging to the additional experiments explained in Section~\ref{sec:discussion}).

\section{Results}
\label{sec:analysis}

We investigate which factors have a significant influence on user performance and on the time they took to complete each trial in the four experiments: \textsc{Test HH}, \textsc{Test HS}, \textsc{Test SH} and \textsc{Test SS}.  The factors we include are: the reference geometry \emph{Gref}, the candidate geometry \emph{Gcand}, and the illumination of the reference sample \emph{Iref}, as well as their first-order interactions (recall that the illumination of the candidate samples remains constant in these behavioral experiments). We also include the \emph{Order} of appearance of each trial. 
We use a general linear mixed model with a binomial distribution for the performance since it is well-suited for binary dependent variables like ours, and a negative binomial distribution for the time, which provides more accurate models than the Poisson distribution by allowing the mean and variance to be different. Since we cannot assume that our observations are independent, we model the potential effect of each particular subject viewing the stimuli as a random effect.
Since we have categorical variables among our predictors, we re-code them to dummy variables for the regression. In all our tests, we fix a significance value ($p$-value) of $0.05$. Finally, for factors that present a significant influence, we further perform pairwise comparisons for all their levels (least significant difference pairwise multiple comparison test).

\subsection{Analysis of user performance and time}
In our online behavioral experiments, we rely on the top-5 accuracy to measure user performance. This metric considers an answer as correct if the reference is among the five candidate materials that the user picked in the trial. Since participants picked five materials ranked in descending order of confidence, the top-1 accuracy could also be considered for our analysis. However, the task they have to solve is not easy and users have an overall top-1 accuracy of 9.21\% which yields sparse results. A random selection would yield a top-1 accuracy of 1\% and a top-5 accuracy of 5\%.

\begin{figure}[t]
	\centering
	\includegraphics[width=\linewidth]{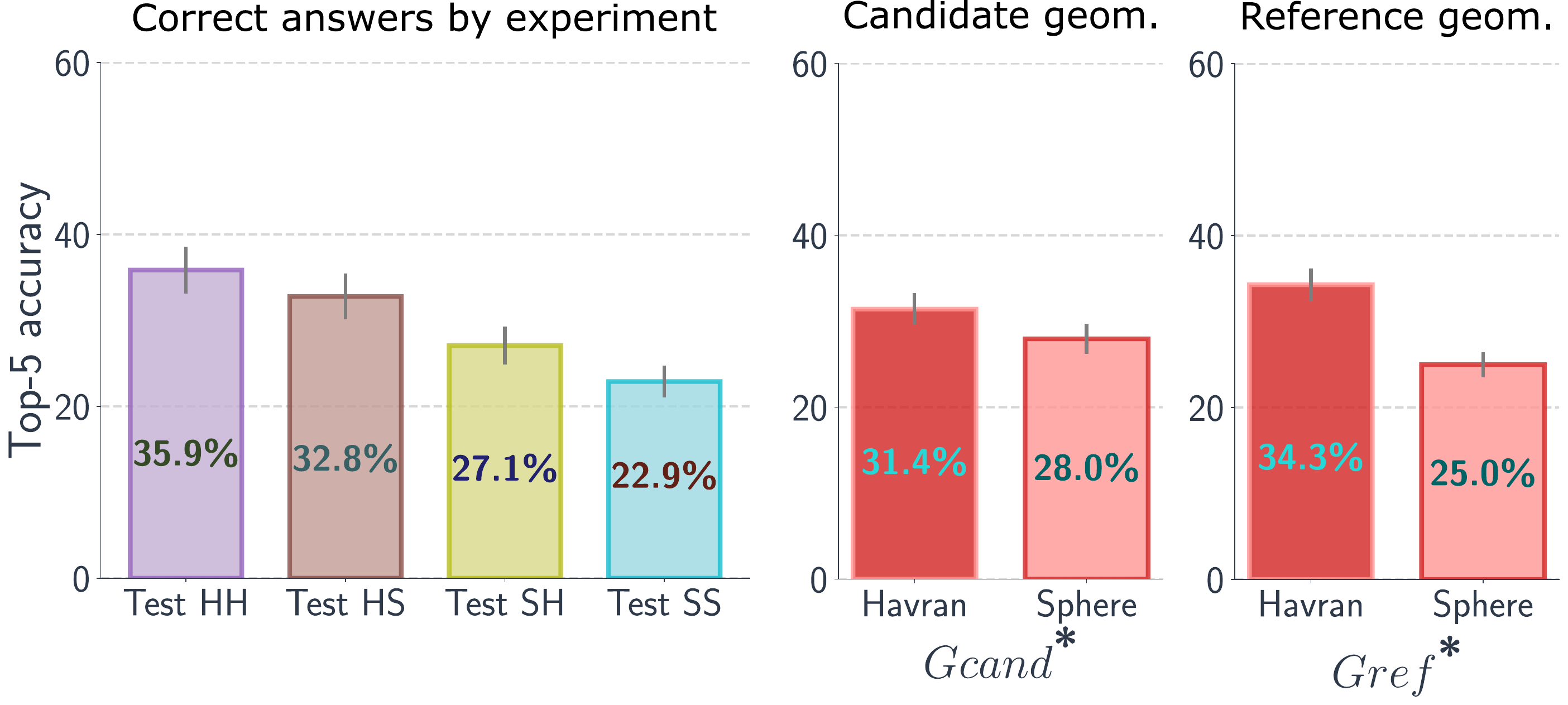}
	\caption{\emph{Left}: Top-5 accuracy for each of the four behavioral experiment. \emph{Center}: Top-5 accuracy for each reference geometry \emph{Gref}. \emph{Right:} Top-5 accuracy for the candidate geometry \emph{Gcand}. We can see how users seem to perform better when the candidate and reference are a high-frequency geometry. All plots have a 95\% confidence interval. The names marked with $*$ are found to have statistically significant differences. }
	\label{fig:top5_gref_gcand}
\end{figure}

\paragraph{Influence of the geometry}
There is a clear effect in user performance when the the geometry changes, regardless if that change happens in the candidate (\emph{Gcand}, $p=0.005$) or the reference geometry (\emph{Gref}, $p<0.001$). This is expected, since the geometry plays a key role in how a surface reflects the incoming light and, therefore, will have an impact on the final appearance of the material. Figure~\ref{fig:top5_gref_gcand} shows user performance in terms of top-5 accuracy with a 95\% confidence interval when the reference and candidate geometry change jointly (left) or individually (center and right). Users seem to perform better when they have to recognize the material in a high-frequency geometry compared to a low-frequency one. 
Those results also suggest that changes in the frequencies of the reference geometry may have a bigger impact on user performance than changes in the frequencies of the candidate geometry (i.e., users perform better with a high-frequency reference geometry and low-frequency candidate geometry, compared to a low-frequency reference geometry and a high-frequency candidate geometry). 

\begin{figure*}[t]
	\centering
	\includegraphics[width=\textwidth]{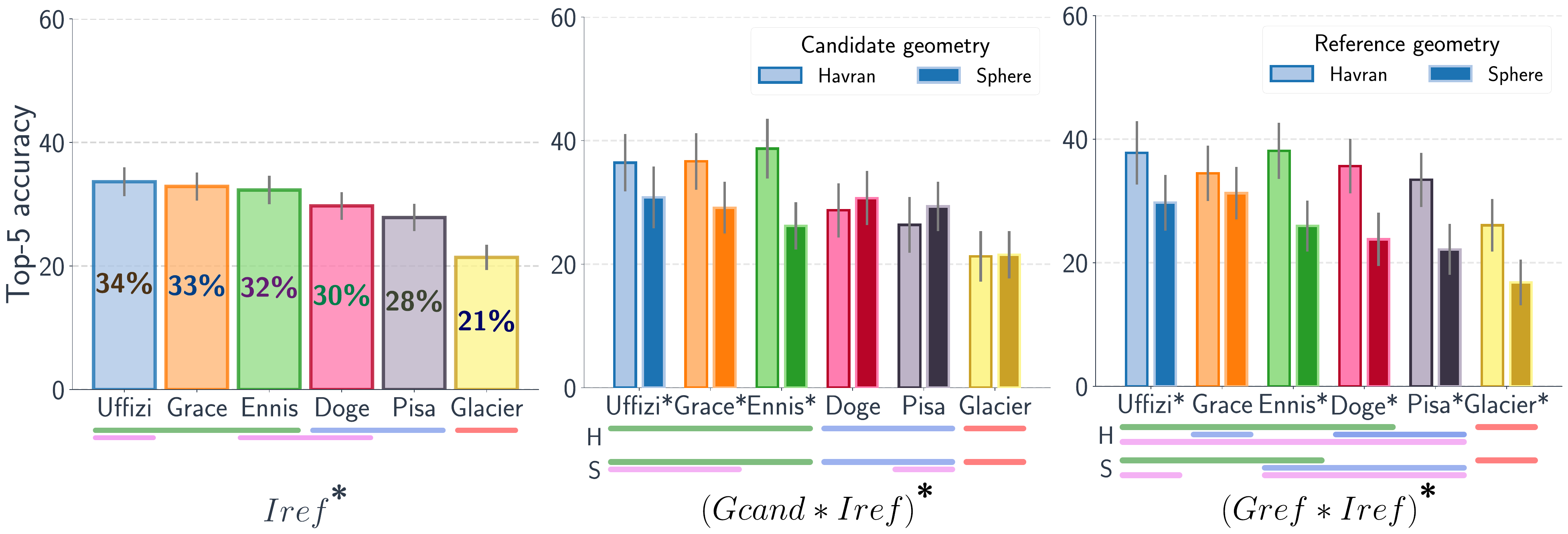}
	\caption{
		\emph{Left: } Top-5 accuracy for each reference illumination (\emph{Iref}). We can see how users seem to perform better with high-frequency illuminations (\emph{Uffizi}, \emph{Grace}, \emph{Ennis}), while their performance is worse with a low-frequency illumination (\emph{Glacier}). Additionally, they have an intermediate performance for medium-frequency illuminations (\emph{Doge} and \emph{Pisa}).
		\emph{Center:} Top-5 accuracy for each reference illumination when the candidate geometry (\emph{Gcand}) changes. We can observe how users appear to perform significantly better with a high-frequency geometry (\emph{Havran}) and illumination. On the other hand, for low-frequency illuminations, changes in the candidate geometry yield statistically indistinguishable performance.	
		\emph{Right:} Top-5 accuracy for each reference illumination when the reference geometry (\emph{Gref}) changes. We can observe how users seem to perform significantly better for all high-frequency illuminations, except for \emph{Grace}.
		The horizontal lines under the x-axis represent groups of statistically indistinguishable performance. We can observe how the groups usually cluster high-, medium- and low-frequency illuminations.
		The reference illuminations marked with $*$ denote significant differences in user performance between geometries for that illumination. The error bars correspond to a 95\% confidence interval.
	}
	\label{fig:top5_by_lref}
\end{figure*}

\paragraph{Influence of the reference illumination}
We observe that the illumination of the reference image has a significant effect on user performance (\emph{Iref}, $p<0.001$). 
This is expected since all the materials in a scene are reflecting the light that reaches them, therefore changes in illumination can significantly influence the final appearance of a material, and how we perceive it~\cite{bousseau2011optimizing}.
Figure~\ref{fig:top5_by_lref}, left, shows the top-5 accuracy for each reference illumination and groups of illuminations with statistically indistinguishable performance. We can observe how users seem to have better performance when the surface they are evaluating has been lit with a high-frequency illumination (\emph{Ennis}, \emph{Grace}, and \emph{Uffizi}); while users appear to perform worse in scenes with a low-frequency illumination (\emph{Glacier}); users show an intermediate performance with a medium-frequency illumination (\emph{Doge} and \emph{Pisa}). Moreover, we performed a least significant difference pairwise multiple comparison test to obtain groups of illuminations with statistically indistinguishable performance. These groups can be observed in Figure~\ref{fig:top5_by_lref}, under the x-axis. If we focus on \emph{Iref} we can see how high- (green), medium- (blue), and low-frequency (red) illuminations yield groups of similar performance. There is an additional group of statistically indistinguishable performance represented in pink color.

\paragraph{Influence of trial order} The order of appearance of the trials during the experiment does not have a significant influence in users performance (\emph{Order}, $p = 0.391$).

\paragraph{First order interactions}
\label{sec:firstorder}
We find that the interaction between the candidate geometry and the reference illumination has a significant effect on user performance ($\text{\emph{Gcand}}*\text{\emph{Iref}}$, $p<0.001$). Users seem to perform better with a high-frequency geometry (compared to a low-frequency one)  when the reference stimuli features a high-frequency illumination (\emph{Iref}=\emph{Uffizi}, $p=0.019$; \emph{Iref}=[\emph{Grace, Ennis}], $p<0.001$). On the other hand, there appears to be no significant changes in performance between a high- and low- frequency candidate geometry when the reference stimuli has a medium- or low-frequency illumination (\emph{Iref}=\emph{Doge}, $p=0.453$; \emph{Iref}=\emph{Pisa}, $p=0.381$; \emph{Iref}=\emph{Glacier}, $p=0.770$). We argue that user performance is driven by the reference sample. When the reference material is lit with a low-frequency illumination, users seem to not be able to properly recognize it. Therefore, changes in the candidate geometry are not relevant to user performance. These results can be seen in Figure~\ref{fig:top5_by_lref}, center. Furthermore, under the x-axis, we can observe the groups with statistically indistinguishable performance where high-, medium-, and low-frequency illuminations yield groups of similar performance.

We also found out that the interaction between the reference geometry and the reference illumination has a significant impact in user performance ($\text{\emph{Gref}}*\text{\emph{Iref}}$, $p=0.012$). Users seem to show better performance for all illuminations with a high-frequency reference geometry (\emph{Gref}=\emph{Havran}, \emph{Iref}=\emph{Uffizi}, $p=0.002$; \emph{Iref} =[\emph{Ennis, Pisa, Doge, Glacier}], $p<0.001$), except for Grace illumination ($p=0.176$), where the differences in humans performance are statistically indistinguishable. 
These results, together with the groups of statistically indistinguishable performance, can be seen in Figure~\ref{fig:top5_by_lref}, right.

In general, we can not conclude that there are significant changes in performance due to the interaction between the candidate and reference geometry (${\text{\emph{Gref}} * \text{\emph{Gcand}}}$, $p = 0.407$). Nevertheless, with a low-frequency reference geometry (\emph{Gref}=\emph{sphere}), users seem to perform significantly better with a high-frequency candidate geometry (\emph{Gcand}=\emph{Havran}, $p=0.009$).

\subsubsection{Analysis of the time spent on each trial}
To account for time, we measure the number of milliseconds that passed since the trial loaded in their screen and until they picked all five materials and pressed the "Continue" button.

\paragraph{Influence of trial order}
We find that the \emph{order} of the trials has a significant influence on the average time users spend to answer them ($p<0.001$). Users spend more time in the first questions and that after few trials the average time they spend becomes stable at around 20 seconds per trial (recall that the \emph{order} does not influence performance). This is expected as users have to familiarize with the experiment during the first iterations. As the test advances, they learn how to interact with it and the time they spend becomes stable. Additional figures and results on the factors that influence the spent time can be found in Appendix~\ref{apx:inf_time}.

\subsection{High-level factors driving material recognition}
In addition to the analysis, we also try to gain intuition on which high-level factors 
drive material recognition, investigate how simple image statistics and image histograms correlate with human answers, and analyze highly non-linear statistics in material classification tasks by training a deep neural network.

\paragraph{Visualizing user answers}
To gain intuition on which high-level factors humans might use while recognizing materials, we employ a stochastic triplet embedding method called t-STE (t-Student stochastic triplet embedding)~\cite{van2012stochastic} directly on user answers. This method maps user answers from their original non-numerical domain into a two-dimensional space that can be easily visualized (find additional details in the Appendix~\ref {apx:tste}). 
Figure~\ref{fig:tste} shows the two-dimensional embeddings after applying the t-STE algorithm to the answers of each online behavioral experiment. Each point in the embedding represents one of the 100 materials from the Lagunas et al. dataset. The insets show the color of each material based on the color classification proposed by Lagunas et al. 
We can observe how materials are clustered by color and, if we focus in a single color, they seem to be clustered by reflectance properties (for instance, in \textsc{Test HH}, red color cluster, we can observe how on the left there are specular materials while on the right there are diffuse materials). This suggests that users have followed a two-step strategy to recognize the materials, and that the high-level factors driving material recognition might be color first, and the reflectance properties second. 
At the end of the \emph{HIT}, users were asked to write the main visual features they used to recognize materials. Out of 368 unique users from the experiments analyzed in Section~\ref{sec:analysis}, 273 answered that they have used the \emph{colors}, and 221 answered that they relied on the \emph{reflections}. Among them, 157 answered both \emph{color} and \emph{reflections} as some of the visual cues they have used to perform the task. This observation, together with the t-STE visualization, strengthens the hypothesis of a two-step strategy.
\begin{figure*}[t]
    \centering
    \includegraphics[width=\textwidth]{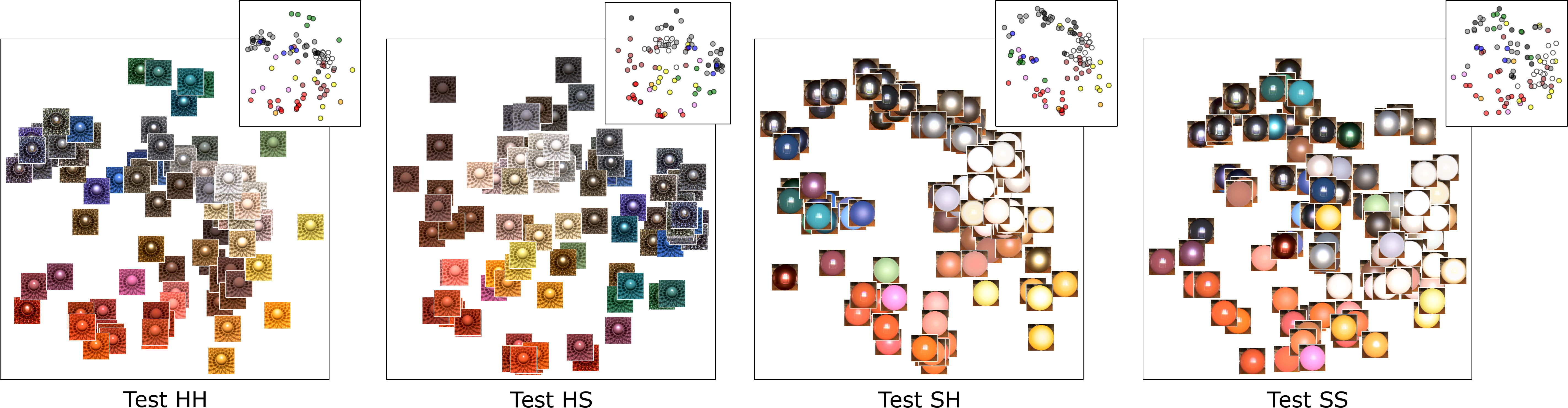}
    \caption{Visualizations of user answers to each of the four online behavioral experiments (namely \textsc{Test HH}, \textsc{Test HS}, \textsc{Test SH}, and \textsc{Test SS}) using the t-STE algorithm~\cite{van2012stochastic}.  The inset shows the color of each material based on the color classification proposed by Lagunas et al.~\shortcite{lagunas2019similarity}. We can see how, for all experiments, materials with similar color properties are grouped together. Furthermore, if we explore the color clusters individually, we can see how there is a second-level arrangement by reflectance properties. These observations suggest that users may be performing a two-step process while recognizing materials where first, they sort them out by color, and second, by reflectance properties.}
    \label{fig:tste}
\end{figure*}

\paragraph{Image statistics} Previous studies focused on simple image statistics as an attempt to further understand our visual system~\cite{motoyoshi2007image,adelson2008image}. Nevertheless, it is argued if our visual system actually derives any aspects of material perception using such simple statistics~\cite{anderson2009image, kim2010image, olkkonen2010perceived}. 
We tested out the correlation between the first four statistical moments of the luminance (considered as the ratio: $L = 0.3086 * R + 0.6094 * G + 0.0820 * B$), the pixel intensity for each color channel independently, and the joint RGB pixel intensity, directly against users top-5 accuracy. To measure correlation we employ a Pearson $\mathcal P$ and Spearman $\mathcal S$ correlation test. We found out that there is little to no correlation except for the standard deviation of the joint RGB pixel intensity where $\mathcal P^2=0.43$ ($p<0.001$) and $\mathcal S^2=0.50$ ($p<0.001$). Additional information can be found in Appendix~\ref{apx:add_statistics}. 

\paragraph{Image histograms}
We also compute the histograms of the RGB pixel intensity, of the luminance, of a Gaussian pyramid~\cite{lee2016fusion}, of a Laplacian pyramid~\cite{burt1983laplacian}, and of log-Gabor filters designed to simulate the receptive field of the simple cells of the Primary Visual Cortex (V1)~\cite{fischer2007self}. To see how such histograms would perform at classifying materials, we train a support vector machine (SVM) that takes the image histogram as the input and classifies the material in that image. 
We use a radial basis function kernel (or Gaussian kernel) in the SVM. We use all image histograms that do not feature \textit{Havran} geometry as the training set and leave the ones with \textit{Havran} as test set. In the end, the best performing SVM uses the RGB image histogram as the input and achieves a 24.17\% top-5 accuracy in the test set.

In addition, we compare the predictions of each SVM directly against human answers. For each reference stimuli we compare the five selections of the user against the five most-likely SVM material predictions for that stimuli. The best SVM uses the histograms of V1-like subband filters and agrees with humans 6.36\% of the time. Moreover, we compare histogram similarities against human answers using a chi-square histogram distance~\cite{pele2010quadratic}. For a reference image stimuli we measure its similarity against all possible candidate image stimuli and compare the closest five against participants’ answers.  The Gaussian pyramid histogram obtained the best result, agreeing with humans 6.29\% of the time. These results show how simple statistics, and higher-order image histograms seem not to be capable of fully capturing human behavior.
We have added additional results on the SVMs and human agreement in Appendix~\ref{apx:add_statistics}.

\paragraph{Image frequencies}
To understand if humans performance could be explained by taking into account the spatial frequency of the reference stimuli, at their viewed size, we have added the high-frequency content measure (HFC), and the first four statistical moments of the reference stimuli magnitude spectrum to the factors analyzed in Section~\ref{sec:analysis}. We found that the Skewness ($p<0.001$) and Kurtosis ($p<0.001$) of the magnitude spectrum seem to have a significant influence on humans performance; however, they present a very small effect size.

\begin{figure}[t]
    \centering
    \includegraphics[width=\linewidth]{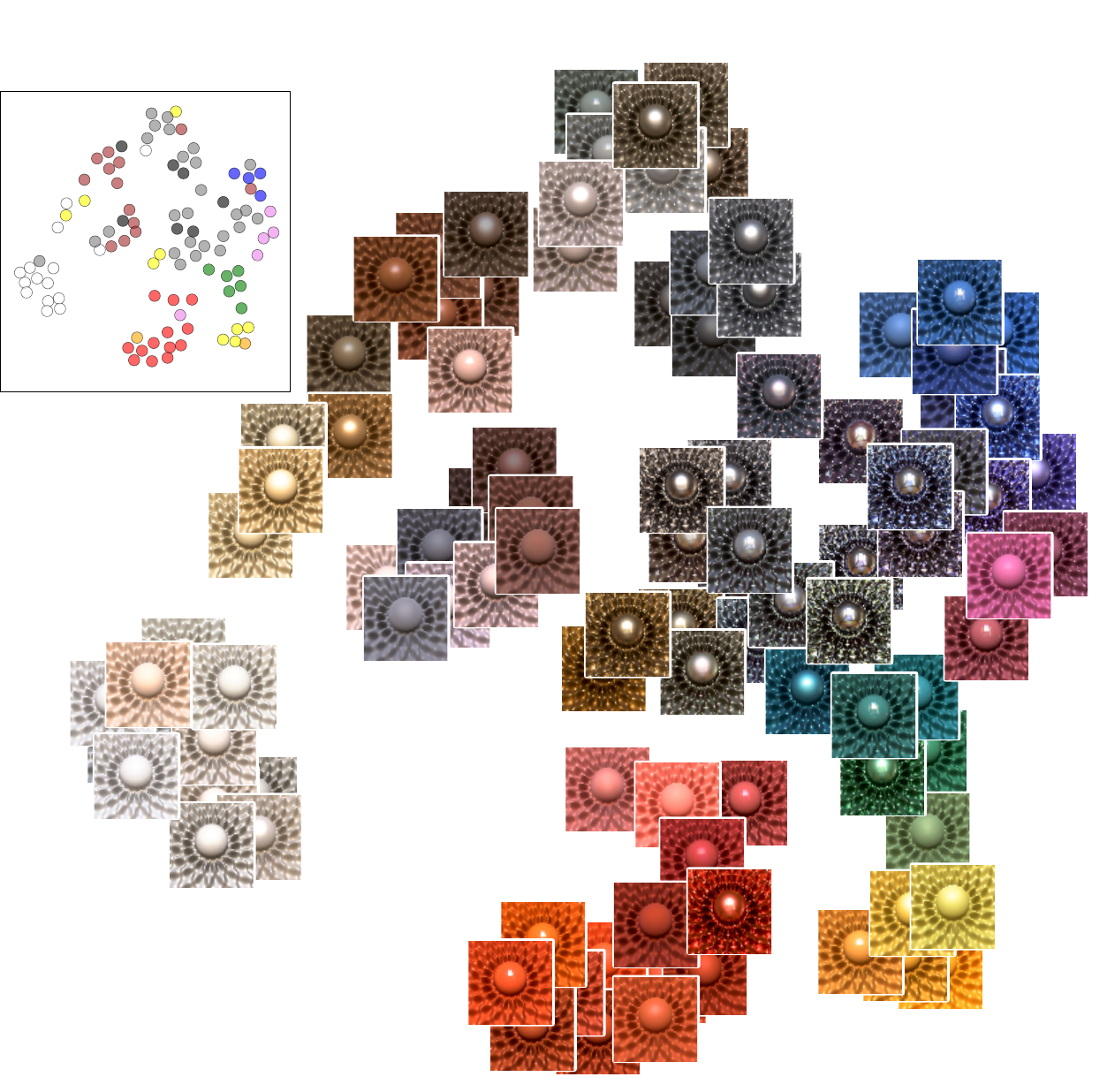}
    \caption{Two-dimensional embedding obtained using the UMAP algorithm~\cite{mcinnes2018umap} on the 128 features of the last fully-connected layer of a ResNet model~\cite{he2016deep} trained to classify materials. The inset shows the color of each material based on the color classification proposed by Lagunas et al.~\shortcite{lagunas2019similarity}. We can observe how materials are arranged by color clusters. Moreover, we can observe similarities between this visualization and the t-STE visualization on user answers.}
    \label{fig:umap}
\end{figure}

\paragraph{Highly non-linear models} Recent studies suggest that, to understand what surrounds us, our visual system is doing an efficient non-linear encoding of the proximal stimulus (the image input to our visual system) and that highly non-linear models might be able to better capture human perception~\cite{fleming2019learning,delanoy2020role}. 
Inspired by this hypothesis, we have trained a deep neural network called ResNet~\cite{he2016deep} employing a loss function suitable to classify the materials in the Lagunas et al. dataset. The images feature the same illuminations as the reference stimuli. We left out the images rendered with \emph{Havran} geometries for validation and testing purposes, and use the rest during training. To know which material the network classifies we add a softmax layer at the end of the network. The softmax layer outputs the probability of the input image to belong to each material in the dataset. In comparison, the model used by Lagunas et al. does not have the last fully-connected and softmax layer, and it is trained using a triplet loss function aiming for similarity instead of classification.
At the end of the training, the model achieves a top-5 accuracy of 89.63\% on the test set, suggesting that such models are actually capable of extracting meaningful features from labeled proximal image data (additional details on the training can be found in Appendix~\ref{apx:resnet}).
To gain intuition on how the network has learned, we have used the UMAP (Uniform Manifold Approximation and Projection) algorithm~\cite{mcinnes2018umap}. This algorithm aims to reduce the dimensionality of a set of feature vectors while maintaining the global and local structure of their original manifold.
Figure~\ref{fig:umap} shows a two-dimensional visualization of the test set obtained using the 
128 features of the fully-connected layer before softmax.
We can observe how materials seem to be grouped first by color and then by its reflectance properties suggesting that the model may have used similar high-level factors to humans when classifying materials.

\begin{figure}[t]
    \centering
    \includegraphics[width=0.8\linewidth]{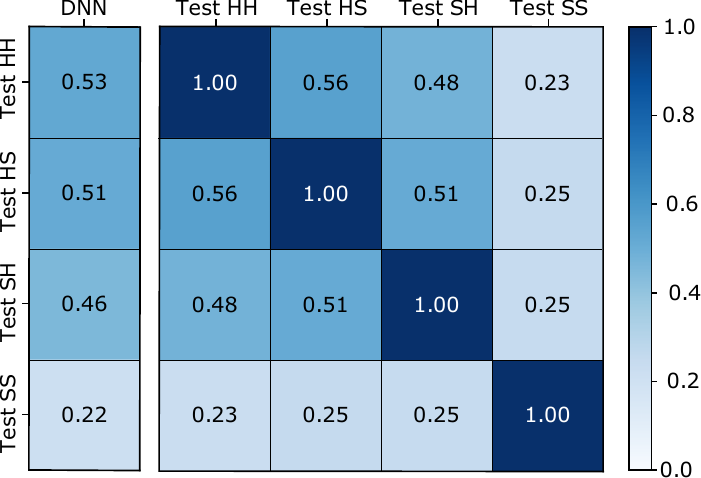}
    \caption{Normalized pairwise similarity for each online behavioral experiment and the deep neural network trained for material classification. We can observe how the pairwise similarity decreases as the stimuli in the experiments cover fewer frequencies in the spectrum, where \emph{Test SS} has the lowest similarity. Additionally, we can see how the similarities between the neural network and each behavioral experiment are on par with those obtained by humans between \emph{Test HH}, \emph{Test HS}, and \emph{Test SH}. }
    \label{fig:sim}
\end{figure}

We additionally assess the degree of similarity between the high-level visualization of each online behavioral experiment and the high-level visualization of the deep neural network. We calculate the similarity in a pairwise fashion where we choose a material sample and retrieve its five nearest neighbors in two different low-dimensional representations. Then, we compute the percentage of materials that are the same in both groups of nearest neighbors. We repeat this process for all the materials and calculate the similarity as the average. The low-dimensional representations are obtained with stochastic methods, where the same input can have different results if we vary the parameters. To evaluate the degree of self-similarity, we run the t-STE algorithm~\cite{van2012stochastic} on each behavioral experiment using five different sets of fully randomly sampled parameters. We obtain a self-similarity value of 0.66, on average across experiments. On the other hand, a set of random low-dimensional representations has a similarity of 0.06, on average. Figure~\ref{fig:sim} shows the average pairwise similarity normalized by the value of self-similarity and random similarity for all experiments and the deep neural network visualization. If we compare between behavioral experiments, we can observe a decreasing degree of similarity as their stimuli feature fewer frequencies in the spectrum, where \textsc{Test SS} yields the lowest similarity in each of the pairwise comparisons. We argue that \textsc{Test SS} has the lowest similarity because it is the experiment where users have the worst performance, thus yielding a \textit{blurry high level visualization}. On the other hand, the network is very accurate at classifying materials and yields a high-level visualization with well-defined material clusters. Moreover, if we focus on the deep neural network visualization, we can observe how its similarity values are, in general, on par with those obtained by users in \textsc{Test HH}, \textsc{Test HS}, and \textsc{Test SH}. This result further supports the hypothesis that both humans and deep neural networks may rely on similar high-level visual features for material recognition tasks. However, this is just a preliminary result that may highlight a future avenue of research, and a thorough analysis of the perceptual relationship between deep learning architectures and humans is out of the scope of this paper.

\section{Discussion}
\label{sec:discussion}

From our online behavioral experiments, we have observed that humans appear to perform better at recognizing materials in stimuli with high-frequency illumination and geometry. 
Moreover, our performance when recognizing materials is poor on low-frequency illuminations, and it remains statistically indistinguishable irrespective of the spatial frequency content in the candidate geometry. 

\begin{figure}[t]
	\centering
	\includegraphics[width=0.9\linewidth]{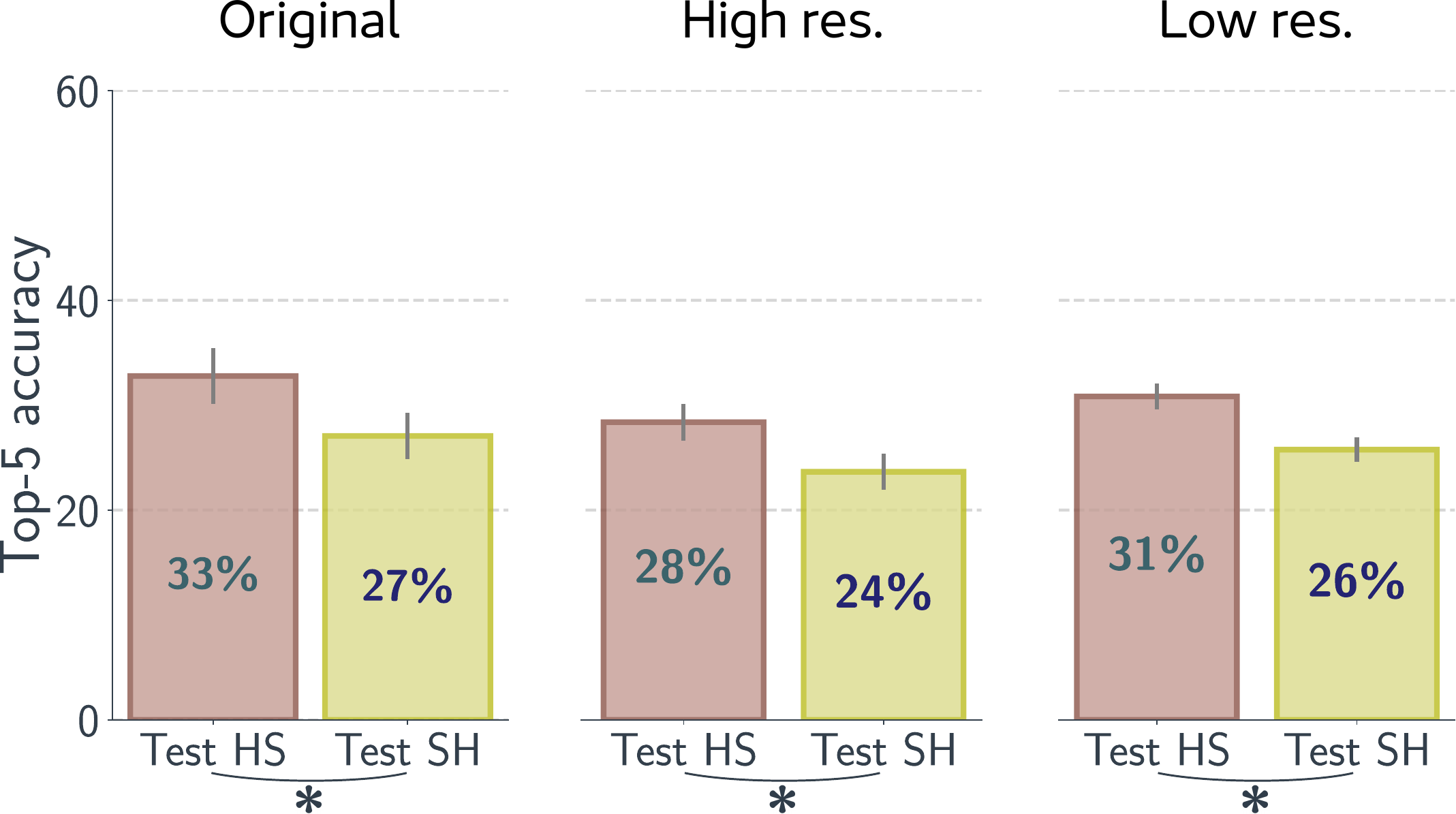}
	\caption{Top-5 accuracy obtained by participants in the original experiment (left), when the stimuli are displayed at 300x300 px. (middle), and at 120x120 px. (right). We can observe how the asymmetric effect of participants performing better when \emph{Havran} is the reference geometry (\textsc{TestHS}) compared to when it is the candidate (\textsc{TestSH}) remains present when the participants observe the reference and candidate stimuli at identical sizes (middle and right). The $*$ denotes significant differences. The error bars correspond to a 95\% confidence interval.}
	\label{fig:assymetric_effect}
\end{figure}

\paragraph{Asymmetric effect of the reference and candidate geometry}
It is also interesting to observe that humans seem to have better performance with a high-frequency reference geometry, compared to a high-frequency candidate geometry ($p=0.001$, see Figure~\ref{fig:assymetric_effect}, left). 
The number of candidates with respect to the reference could be used as an explanation for this observation, since users may devote more time to inspecting the single reference than the higher number of candidates. 
At the same time, a lower performance with a high-frequency reference geometry may speak against an inverse optics approach since having multiple candidate materials with the same geometry and illumination could provide a strong cue to inferring the material.

One potential factor that may explain this difference in performance is the different display sizes of the reference (300x300 px.) and the candidate (120x120 px.) stimuli. To test this hypothesis, we have launched two additional experiments where we collect answers on \textsc{Test HS} and \textsc{Test SH} displaying the candidate and the reference stimuli at size 300x300, and other two additional experiments where they are displayed at 120x120 px. We sample the stimuli to cover all the possible combinations of illuminations and materials and keep other technical details as explained in Section~\ref{sec:user study}. 
We perform an analysis of the gathered data similar to the one explained in Section \ref{sec:analysis}, but using the different experiment type as a factor.
From our results we observe that such asymmetric effect remains present when the stimuli are displayed at 300x300 px. ($p<0.001$) and when they are displayed at 120x120 px. ($p<0.001$). Those results can be seen in Figure~\ref{fig:assymetric_effect}, middle and right.
It is also interesting to observe how users have slightly worse performance when the stimuli are displayed at 300x300 px. At such display size only three candidate stimuli per row could be displayed taking into account the most used display size. Thus, seems reasonable to think that the need for additional scrolling could be hampering participants performance.

\begin{figure}[t]
	\centering
	\includegraphics[width=\linewidth]{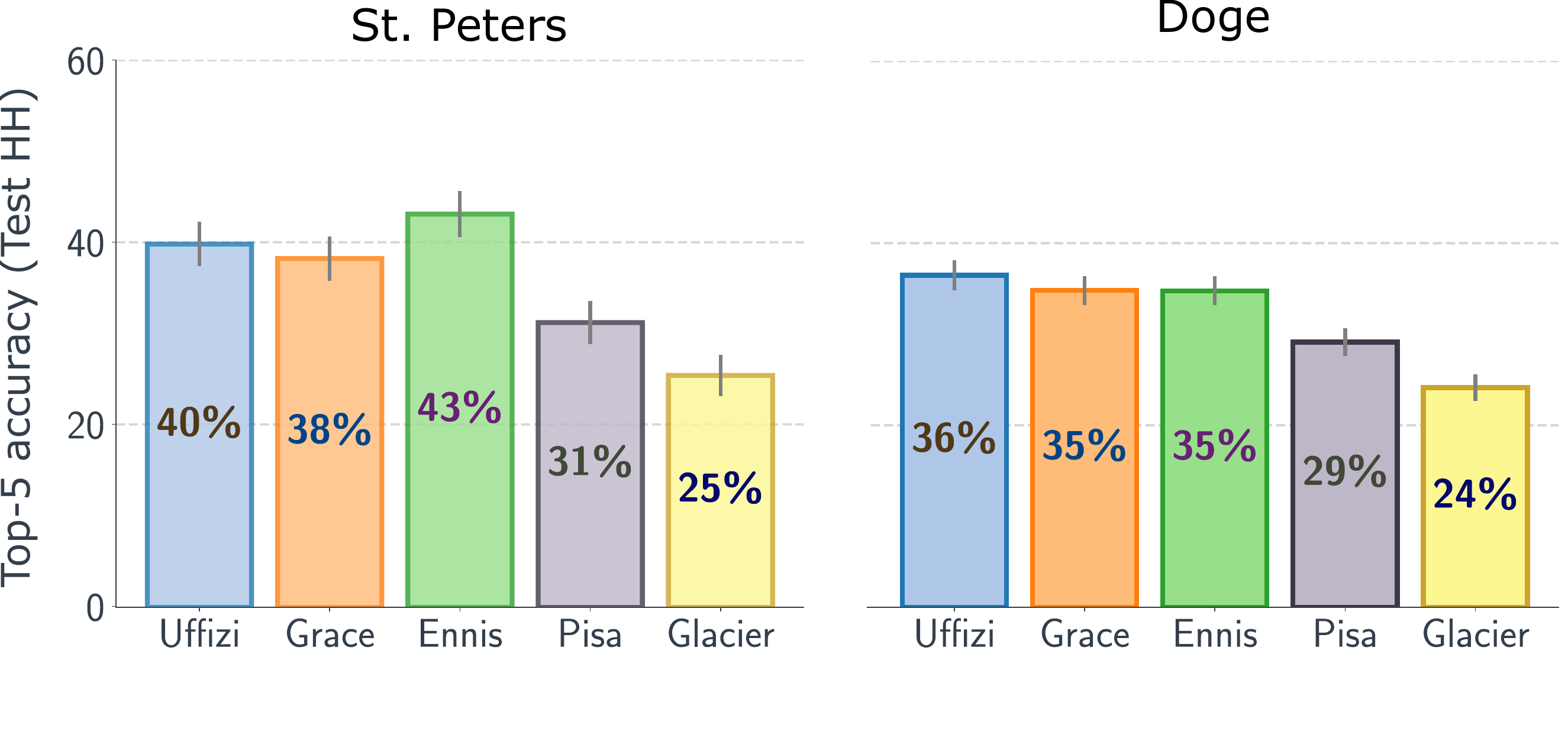}
	\caption{\emph{Left:} Top-5 accuracy for each reference illumination when \emph{St. Peters}, a high-frequency illumination, is the candidate illumination. \emph{Right}: Top-5 accuracy for each reference illumination when \emph{Doge}, a medium-frequency illumination, is the candidate illumination. Both results have been obtained for \textsc{Test HH}.
	We can observe how, for both candidate illuminations, participants seem to perform better with high-frequency reference illuminations (\emph{Uffizi}, \emph{Grace}, \emph{Ennis}), they perform worse with medium-frequency ones (\emph{Pisa}), and have their worst performance with low-frequency reference illuminations (\emph{Glacier}). In addition, we also observe that participants have slightly better performance when \emph{St. Peters} (high-frequency illumination) is the candidate illumination. The error bars correspond to a 95\% confidence interval.}
	\label{fig:top5_iref_doge_icand}
\end{figure}

\paragraph{Influence of the candidate illumination}
We have seen that humans seem to be better at recognizing materials under high-frequency reference illuminations. However, in Figure~\ref{fig:ref_illum} and \ref{fig:hfc} we can see that the \emph{St. Peters} candidate illumination features a similar frequency content to the reference illuminations where users have better performance. 
To asses if \emph{St. Peters} illumination contains a set of frequencies that aids recognizing materials under reference illuminations with a similar set of frequencies, we have launched an additional behavioral experiment. 
In this experiment we use \emph{Doge}, a medium-frequency illumination, as the candidate illumination. We sample the stimuli to cover all materials and reference illuminations in \textsc{Test HH}. Other technical details are kept as explained in Section~\ref{sec:user study}. 
From the data collected (see Figure~\ref{fig:top5_iref_doge_icand}), we can observe how, using \emph{Doge} as the candidate illumination, humans performance follows a similar distribution to the original experiment (with \emph{St. Peters} as the candidate illumination). Participants seem to perform better with high-frequency reference illuminations (\emph{Uffizi}, \emph{Grace}, \emph{Ennis}), they perform worse with medium-frequency ones (\emph{Pisa}), and have their worst performance with low-frequency reference illuminations (\emph{Glacier}). In addition, participants seem to have slightly better performance with a high-frequency candidate illumination (\emph{St. Peters}) compared to a medium-frequency one (\emph{Doge}).

\paragraph{Interplay between material, geometry, and illumination}
We have looked into how geometry, illumination, and their frequencies affect our performance in material recognition tasks. Our stimuli were rendered images in which we varied the frequency of the illumination, and of the underlying geometry of the object present. To better understand how our factors (illumination and geometry) affect the generated stimuli, and thus the proximal stimulus, we offer here a brief description of the rendering equation, providing an explanation of the probable effect of how the frequencies of the geometry and illumination in the 3D scene affect the final, rendered image that is used as a stimulus in our experiments.
From the point of view of the rendering equation, the radiance $L_o$ at point $x$ in direction $\omega_o$, assuming a distant illumination and non-emissive surfaces can be approximated as
\begin{equation}
    L_o(x, \omega_o) \approx \int_\Omega
    L_i(\omega_i)
    \; F(\omega_i, \omega_o)
    \; T(x, \omega_i, \omega_o)
    \; d\omega_i
    \label{eq:render},
\end{equation}
where $L_i$ accounts for the incoming light, the variable  $F$ accounts for the reflectance of the surface, and $T$ depends on the point of the surface we are evaluating, therefore, on the geometry.

\begin{figure}[t]
    \centering
    \includegraphics[width=0.9\columnwidth]{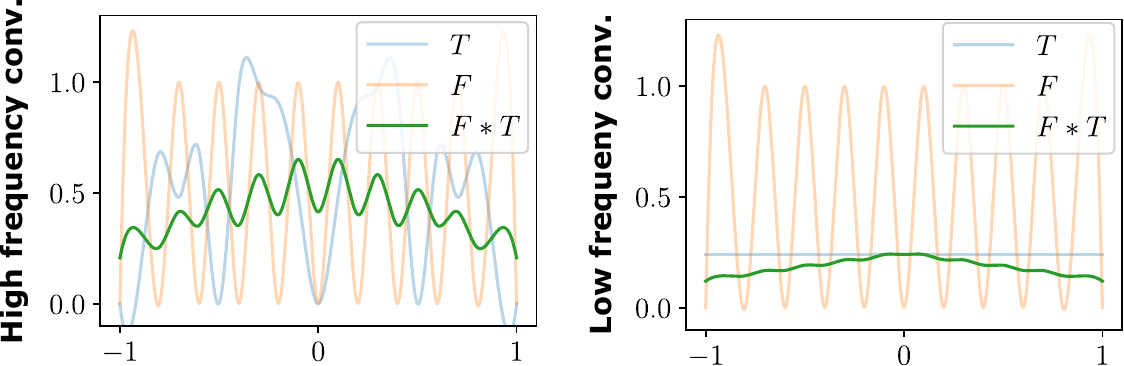}
    \caption{Example of a convolution ($F * T$, green line) between a material ($F$, orange line) and a geometry ($T$, blue line) with different frequency content. \emph{Left:} We can see how when we convolve a geometry and a material with high spatial frequencies, the resulting convolution also retains high-frequency content. \emph{Right:} We observe how when geometry has low spatial frequencies and the material has high spatial frequencies, the resulting convolution does not retain high-frequency content. Note that $T$ and $F$ are not necessarily related to a real BRDF or shape from the ones reported in this manuscript.}
    \label{fig:freq_example}
\end{figure}

The simulation of the radiance $L_o$ can be seen as a convolution (spherical rotation)~\cite{ramamoorthi2001efficient} between each signal: incoming radiance $L_i$, material $F$ and geometry $T$. Moreover, if we analyze $L_o$ in the frequency domain (where $\mathcal F$ is the Fourier transform), and apply the convolution theorem ($f*g = \mathcal F(f)\cdot \mathcal F(g)$) the value of ${\mathcal F \big( L_o \big)}$ becomes
\begin{equation}
    \mathcal F \big(L_o\big) \approx
    \mathcal F \big(L_i\big)
    \; \mathcal F\big(F\big)
    \; \mathcal F\big(T\big)
    \label{eq:render_freq}.
\end{equation}
Equation~\ref{eq:render_freq} shows how the frequency of the radiance $L_o$ in the final image is a multiplication of all the other signals $L_i$, $F$, and $T$ in the frequency domain. Thus, the final image will only have the frequencies that are contained within the three other signals. Figure~\ref{fig:freq_example} shows how when we convolve two high-frequency signals, the resulting one keeps the high-frequency content; on the other hand, when we convolve a high- and a low-frequency signal, the resulting one has most of its frequencies masked.

We can relate the observations made from Equation~\ref{eq:render_freq} to the results on user performance that we obtained from the online behavioral experiments. We have seen that users seem to consistently perform better when they recognize materials from high-frequency geometries and illuminations. This finding is supported by Equation~\ref{eq:render_freq} since, to avoid filtering the frequencies of the material in the stimuli, it should have a high-frequency geometry and illumination. Moreover, a low-frequency geometry (or illumination) could filter out the frequencies of the illumination (or geometry) and the material, thus yielding fewer visual features on the final image and, as a result, worse users performance. This is consistent with our findings from the analysis of first-order interactions for users performance in Section~\ref{sec:firstorder}.

\begin{figure}[t]
    \centering
    \includegraphics[width=\linewidth]{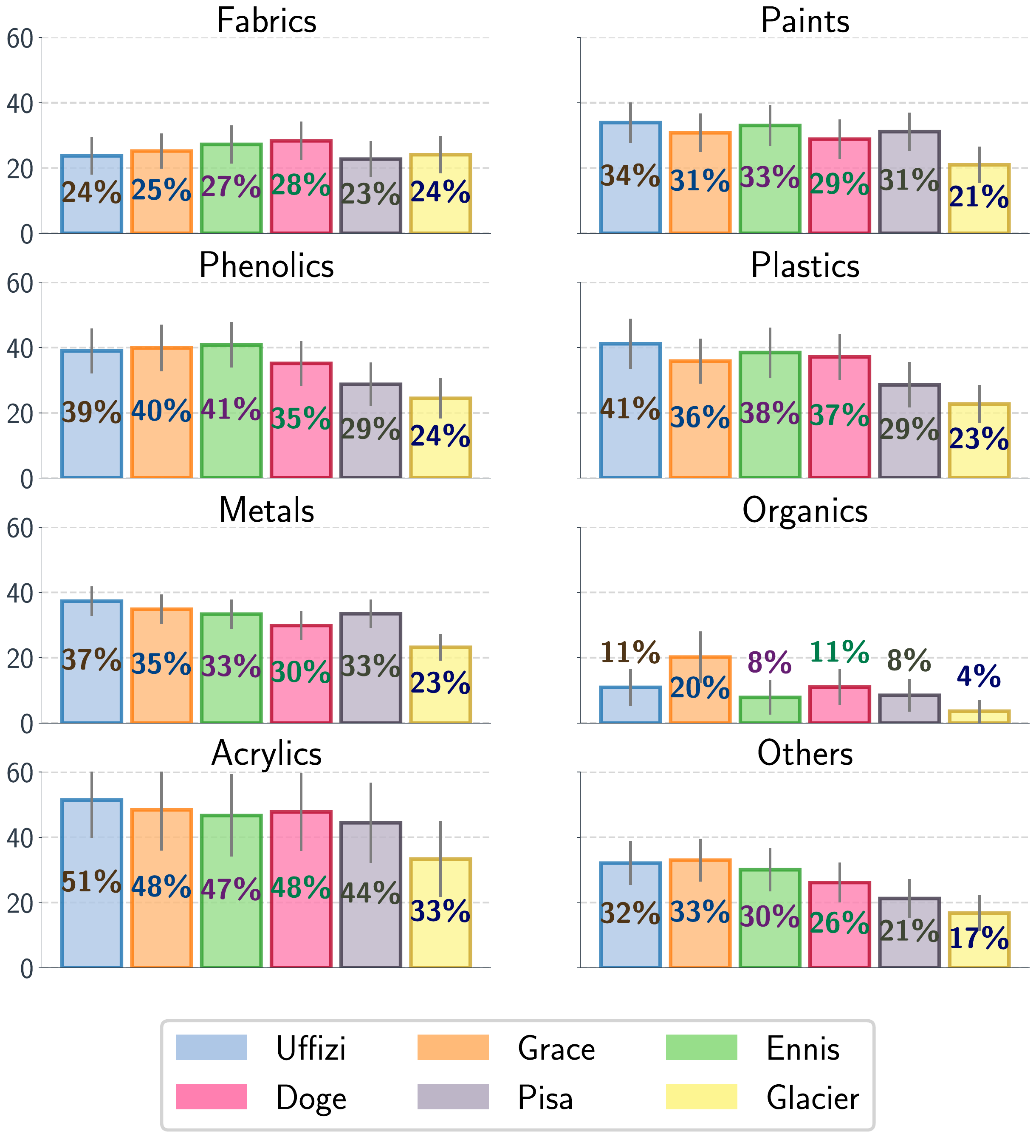}
    \caption{Users performance, in terms of top-5 accuracy, for material recognition tasks taking into account the reflectance of the materials. We can observe how, on average, users perform better for high-frequency illuminations (\emph{Uffizi, Grace, and Ennis}). Also, we can see how for classes, like \emph{fabrics} or \emph{organics}, containing materials with diffuse surface reflectance (low-frequency), users do not have better performance with broad frequency content illuminations. 
    We argue that, since they have a low-frequency surface reflectance, even though there is a high-frequency illumination, those frequencies cannot be represented on the final stimulus that is input to our visual system.}
    \label{fig:top5_lref_and_reflectance}
\end{figure}

\paragraph{Material categories} We have seen that the reflectance properties seem to be one of the main high-level factors driving material recognition. In this regard, we have also investigated users performance using the classification by reflectance type proposed by Lagunas et al.~\shortcite{lagunas2019similarity}, where the MERL database is divided into eight different categories with similar reflectance properties. On average, users perform best on \emph{acrylics}, with a top-5 accuracy of 45.45\%, while they have their worst performance with \emph{organics}, with an accuracy of 10.22\%.
Figure~\ref{fig:top5_lref_and_reflectance} shows the top-5 accuracy for each category in each reference illumination. Firstly, we observe that users seem to perform better with high-frequency illuminations (\emph{Uffizi}, \emph{Grace}, \emph{Ennis}). However, we can see how \emph{fabrics} and \emph{organics} do not follow this trend.
We argue that \emph{fabrics} and \emph{organics} contain mostly materials with a diffuse surface reflectance (low-frequency) that clamp the frequencies of the illumination and, therefore, yield fewer cues in the final stimulus that is input to our visual system. 

\section{Conclusions}

In this work, we have presented a thorough and systematic analysis of the interplay between geometry, illumination, and their spatial frequencies in human performance recognizing materials. We launched rigorous crowd-sourced online behavioral experiments where participants had to solve a material recognition task between a reference and a set of candidate samples.
From our experiments, we have observed that, in general, humans appear to be better at recognizing materials in a high-frequency illumination and geometry. 
We found that simple image statistics, image histograms, and histograms of V1-like subband filters are not capable of capturing human behavior, and, additionally, explored highly non-linear statistics by training a deep neural network on material classification tasks. 
We showed that deep neural networks can accurately perform material classification, which suggests that they are capable of encoding and extracting meaningful information from labeled proximal image data. In addition, we gained intuition on which are the main high-level factors that humans and those highly non-linear statistics use for material recognition and found preliminary evidence that such statistics and humans may share similar high-level factors for material recognition tasks.

\paragraph{Limitations and future work}
To collect data for the online behavioral experiment we have relied on the Lagunas et al.~\cite{lagunas2019similarity} dataset which contains images of a diverse set of materials, geometries, and illuminations that faithfully resemble their real-world counterparts. This database focuses on isotropic materials, which are capable of modeling only a subset of real-world materials. A systematic and comprehensive analysis of other heterogeneous materials, or an extension of this study to other non-photorealistic domains, remains to be done.
Our stimuli were rendered using the sphere and \emph{Havran} geometry, although those surfaces have been widely used in the literature~\cite{lagunas2019similarity,serrano2016,jarabo2014btf,havran2016}, introducing new geometries could help to further analyze the contribution of the spatial frequencies of the geometry in our perception of material appearance~\cite{nishida1998use}.
Moreover, to select our stimuli, we characterized the frequency content of real-world illuminations using the high-frequency content measure~\cite{brossier2004real}. We focus on real-world illuminations, which are by definition broadband, therefore, we do not impose nor limit their frequency distribution in our analyses; carefully controlling the spatial frequency of the stimuli via filtering in order to isolate frequency bands and study their individual contribution to the process of material recognition is an interesting avenue of research.

In our additional experiments, we have investigated the asymmetric effect in performance with a high-frequency reference geometry, compared to a high-frequency candidate geometry when all stimuli are displayed at the same size. A rigorous study of the interplay between display size, the spatial frequencies of the stimuli, and how this affects humans performance on material recognition remains an interesting line of future work.
Furthermore, despite the fact that our neural network was trained to classify materials, without any sort of perceptual information, it achieved an agreement with participants answers of 22.43\%. This does not prove that the neural network follows the same mechanisms as humans do when performing these tasks.  However, this result together with the increase in popularity of deep neural networks, makes the analysis of the perceptual relationship between learned features and the features that our visual system uses to recognize materials a promising avenue to explore.
Last, we hope that our analyses will provide relevant insights that will help shed light on the underlying perceptual processes that occur when we recognize materials and, in particular, on how the confounding factors of geometry and illumination affect our perception of material appearance.

\section*{Acknowledgements}
We want to thank the anonymous reviewers for their encouraging and insightful feedback on the manuscript.  Also, we want to thank Sandra Malpica, Elena Garces, Ibon Guillen, and Adrian Jarabo for the early discussions about the paper, and Dani Martin and Ibon Guillen for their help proofreading. This project has received funding from the European Research Council (ERC) under the European Union’s Horizon 2020 research and innovation programme (CHAMELEON project, grant agreement No 682080), from the European Union’s Horizon 2020 research and innovation programme under the Marie Skłodowska-Curie grant agreements No 765121 and 956585, and from the Spanish Ministry of Economy and Competitiveness (projects TIN2016-78753-P and PID2019-105004GB-I00).

\appendix 

\section*{Appendix}

\section{Additional results on the influence of time}
\label{apx:inf_time}
\begin{figure}[t]
	\centering
	\includegraphics[width=\linewidth]{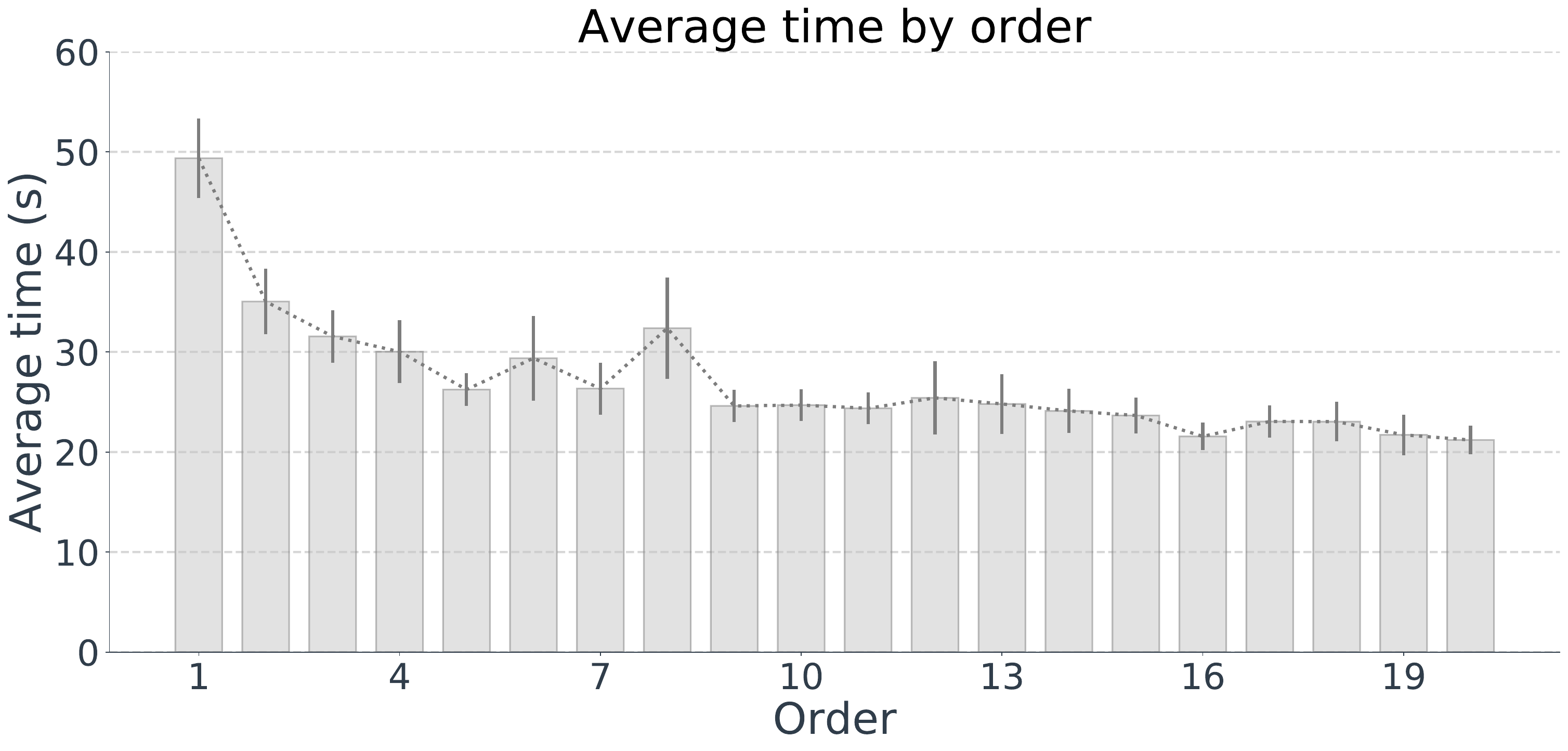}
	\caption{Average time the users spent for each trial according to the order of appearance during the online behavioral experiment. We can observe how, as the user progresses through the experiment, the time spent on each trial becomes stable. 
	The error bars correspond to a 95\% confidence interval.}
	\label{fig:time}
\end{figure}

Additional details on the time that each participant spent doing the online behavioral experiment. In Figure~\ref{fig:time} we can see how the time spent to answer each trial becomes stable as the behavioral experiment advances.

\paragraph{Influence of reference illumination}
The reference illumination \emph{Iref} influences the time users spend to answer each trial ($p=0.001$). Users spend more time when the stimuli are lit with \emph{Ennis} illumination while they are the fastest when the illumination is \emph{Doge}.

We did not find a significant influence of the reference geometry \emph{Gref} or candidate geometry \emph{Gcand} in the average time each user spent to answer each trial.

\paragraph{First order interactions}
We observe that users take significantly longer to answer the trials when both the reference geometry and the candidate geometry change (\emph{Gref} $*$ \emph{Gcand}, $p=0.001$). This happens in the case where the reference geometry has mostly low spatial frequency content and the candidate geometry changes (\emph{Gcand}=\emph{sphere}, $p$=0.002); and when the reference has mostly low spatial frequency (\emph{Gref}=\emph{sphere}, $p$=0.001) and the candidate geometry changes.

\section{Additional details on the  t-STE algorithm}
\label{apx:tste}
The t-STE algorithm aims to obtain an n-dimensional embedding that satisfies as many qualitative comparisons of the type \emph{"A is more similar to B than C"} as possible. In our case, a two-dimensional embedding which is easier to visualize. Nevertheless, in the user studies, we have asked participants to select five materials from a pool of candidates and we do not have such qualitative comparisons. However, we can assume that the selection of the users will be closer (more similar) to the reference than any other material that was not selected. Based on this assumption, we generate triplets where the user selection is more similar to the reference material than any other random material that is not within the 5 selected materials. We repeat this process ten times for each of the 5 materials selected by the user making sure that the new randomly sampled material has not been randomly selected already nor that it belongs to the pool of 5 selected materials.
To run the t-STE we set a learning rate of 1, and an $\alpha=25$ (degrees of freedom of the Student-t kernel). Additionally, we apply a logarithmic transformation to the loss value of the t-STE. Those parameters are the same for the answers of the four experiments.

\section{Additional details on image statistics}
\label{apx:add_statistics}
To measure the correlation between image statistics and users performance we employ a Pearson $\mathcal P$ and Spearman $\mathcal S$ correlation test with a significance value ($p$-value) of 0.05. The value $\mathcal P^n$ represents the Pearson correlation for the $n^{\text{th}}$ statistical moment (same applies for the Spearman $\mathcal S^n$ correlation).

\paragraph{Luminance} We analyze if the moments of the luminance of each material image have a direct influence on users performance. We found that the moments of the luminance are not correlated with users performance:
$\mathcal P^1=-0.14$ ($p=0.17$), $\mathcal S^1=-0.15$ ($p=0.15$), 
$\mathcal P^2=0.02$ ($p=0.83$), $\mathcal S^2=-0.03$ ($p=0.78$), 
$\mathcal P^3=0.03$ ($p=0.77$), $\mathcal S^3=0.03$ ($p=0.78$), 
$\mathcal P^4=0.01$ ($p=0.94$), $\mathcal S^4=0.05$ ($p=0.65$).

\paragraph{RGB image} We analyze if the moments of the joint RGB intensity of each material image have a direct influence on users performance. We found that the moments of the joint RGB intensity have little to no correlation with users performance except for the standard deviation:
$\mathcal P^1=-0.02$ ($p=0.79$), $\mathcal S^1=-0.06$ ($p=0.51$), 
$\mathcal P^2=0.43$ ($p<0.001$), $\mathcal S^2=0.50$ ($p<0.001$), 
$\mathcal P^3=0.16$ ($p=0.09$), $\mathcal S^3=0.22$ ($p=0.02$), 
$\mathcal P^4=-0.1$ ($p=0.30$), $\mathcal S^4=-0.06$ ($p=0.52$).

\bigskip
We also tested out the correlation for each channel and found out that for all the channels there is no correlation for any of the first 4 statistical moments.

\paragraph{Red channel}
On the red channel there seems to be a slight positive linear correlation between the fourth moment (kurtosis) and users performance. All the other statistics show no significant correlation:
$\mathcal P^1=-0.10$ ($p=0.29$), $\mathcal S^1=-0.08$ ($p=0.42$), 
$\mathcal P^2= 0.03$ ($p=0.60$), $\mathcal S^2=-0.02$ ($p=0.87$), 
$\mathcal P^3= 0.07$ ($p=0.46$), $\mathcal S^3=0.07$ ($p=0.51$), 
$\mathcal P^4= 0.20$ ($p=0.04$), $\mathcal S^4=0.15$ ($p=0.13$).

\paragraph{Green channel}
There is no correlation between any statistics on the green channel:
$\mathcal P^1=-0.04$ ($p=0.66$), $\mathcal S^1=-0.0.$ ($p=0.74$), 
$\mathcal P^2= 0.03$ ($p=0.55$), $\mathcal S^2= 0.04$ ($p=0.67$), 
$\mathcal P^3= 0.05$ ($p=0.64$), $\mathcal S^3= 0.06$ ($p=0.53$), 
$\mathcal P^4= 0.05$ ($p=0.63$), $\mathcal S^4= 0.01$ ($p=0.94$).

\paragraph{Blue channel}
Similar to the green channel, the blue does not show any correlation for the first 4 statistical moments:
$\mathcal P^1= 0.03$ ($p=0.72$), $\mathcal S^1=-0.004$ ($p=0.93$), 
$\mathcal P^2= 0.06$ ($p=0.52$), $\mathcal S^2= 0.01$ ($p=0.95$), 
$\mathcal P^3= 0.13$ ($p=0.19$), $\mathcal S^3= 0.10$ ($p=0.30$), 
$\mathcal P^4= 0.16$ ($p=0.11$), $\mathcal S^4=-0.05$ ($p=0.61$).

\subsection{Additional results on the SVMs and histogram similarity}
We have trained a total of 6 support vector machine models, each of them using a different input: RGB pixel intensity, luminance intensity, Gaussian pyramid pixel intensity~\cite{lee2016fusion}, Laplacian pyramid pixel intensity~\cite{burt1983laplacian}, joining the Gaussian and Laplacian pyramids, and using log-Gabor filters~\cite{fischer2007self}. For each of them the SVM achieved a top-5 accuracy in the test set of: 24.17\%, 15.16\%, 22.50\%, 6.33\%, 7.52\%, and 16.33\%, respectively. 
In addition, we have compared how the SVM predictions agreed with humans' answers from the online behavioral experiments. For each SVM the agreement is: 4.24\%, 4.33\%, 4.34\%, 5.04\%, 4.97\%, and 6.36\% respectively.
Last, we have also computed the histogram similarity using a chi-squared distance. Then, we have taken the five closest samples and compared that with human answers. We do that for each of the five different histograms and each achieves an agreement of: 5.95\%, 5.45\%, 6.29\%, 4.97\%, 5.04\%, and 5.07\% respectively.

\section{Additional details on ResNet training}
\label{apx:resnet}
To train the 35 layers ResNet (34 of the original model plus an additional fully connected) ~\cite{he2016deep} we have employed the dataset introduced by Lagunas et al.~\cite{lagunas2019similarity}, which contains renderings of materials with different illuminations and geometries. We keep the images rendered with \emph{Havran-3} geometry for validation purposes and \emph{Havran} geometry for testing.  All the other images are used for training. To train the model to classify materials we use a soft cross-entropy loss where samples that do not belong to the same class are penalized~\cite{szegedy2015}. The loss function takes the probabilities output of the softmax layer and penalizes when they give a high probability to the materials that do not belong to the input image. The images input to the model are resized to 224x224 px.
The parameters of the model are initialized using a pretrained version on ImageNet dataset~\cite{deng2009imagenet}. 
We use the ADAM algorithm~\cite{kingma2014} as the optimizer. The model has been trained during 50 iterations starting at a learning rate of $10^{-3}$ and decayed by a factor of 10 at the iteration 20, 35, and 45; the batch-size was set to 64 images. We use the PyTorch framework and use an Nvidia 2080Ti GPU. 

\end{sloppypar}

\bibliographystyle{ACM-Reference-Format}
\bibliography{paper}


\begin{thebibliography}{83}


\ifx \showCODEN    \undefined \def \showCODEN     #1{\unskip}     \fi
\ifx \showDOI      \undefined \def \showDOI       #1{#1}\fi
\ifx \showISBNx    \undefined \def \showISBNx     #1{\unskip}     \fi
\ifx \showISBNxiii \undefined \def \showISBNxiii  #1{\unskip}     \fi
\ifx \showISSN     \undefined \def \showISSN      #1{\unskip}     \fi
\ifx \showLCCN     \undefined \def \showLCCN      #1{\unskip}     \fi
\ifx \shownote     \undefined \def \shownote      #1{#1}          \fi
\ifx \showarticletitle \undefined \def \showarticletitle #1{#1}   \fi
\ifx \showURL      \undefined \def \showURL       {\relax}        \fi
\providecommand\bibfield[2]{#2}
\providecommand\bibinfo[2]{#2}
\providecommand\natexlab[1]{#1}
\providecommand\showeprint[2][]{arXiv:#2}

\bibitem[\protect\citeauthoryear{Adelson}{Adelson}{2000}]%
        {adelson2000lightness}
\bibfield{author}{\bibinfo{person}{EH Adelson}.}
  \bibinfo{year}{2000}\natexlab{}.
\newblock \bibinfo{title}{Lightness perception and lightness illusions. The new
  cognitive neurosciences}.
\newblock
\newblock


\bibitem[\protect\citeauthoryear{Adelson}{Adelson}{2001}]%
        {adelson2001seeing}
\bibfield{author}{\bibinfo{person}{Edward~H Adelson}.}
  \bibinfo{year}{2001}\natexlab{}.
\newblock \showarticletitle{On seeing stuff: the perception of materials by
  humans and machines}. In \bibinfo{booktitle}{\emph{Human Vision and
  Electronic Imaging VI}}, Vol.~\bibinfo{volume}{4299}. International Society
  for Optics and Photonics, \bibinfo{pages}{1--13}.
\newblock


\bibitem[\protect\citeauthoryear{Adelson}{Adelson}{2008}]%
        {adelson2008image}
\bibfield{author}{\bibinfo{person}{Edward~H Adelson}.}
  \bibinfo{year}{2008}\natexlab{}.
\newblock \showarticletitle{Image statistics and surface perception}. In
  \bibinfo{booktitle}{\emph{Human Vision and Electronic Imaging XIII}},
  Vol.~\bibinfo{volume}{6806}. International Society for Optics and Photonics,
  \bibinfo{pages}{680602}.
\newblock


\bibitem[\protect\citeauthoryear{Anderson}{Anderson}{2011}]%
        {anderson2011}
\bibfield{author}{\bibinfo{person}{Barton~L Anderson}.}
  \bibinfo{year}{2011}\natexlab{}.
\newblock \showarticletitle{Visual perception of materials and surfaces}.
\newblock \bibinfo{journal}{\emph{Current Biology}} \bibinfo{volume}{21},
  \bibinfo{number}{24} (\bibinfo{year}{2011}), \bibinfo{pages}{R978--R983}.
\newblock


\bibitem[\protect\citeauthoryear{Anderson and Kim}{Anderson and Kim}{2009}]%
        {anderson2009image}
\bibfield{author}{\bibinfo{person}{Barton~L Anderson} {and}
  \bibinfo{person}{Juno Kim}.} \bibinfo{year}{2009}\natexlab{}.
\newblock \showarticletitle{Image statistics do not explain the perception of
  gloss and lightness}.
\newblock \bibinfo{journal}{\emph{Journal of Vision (JOV)}}
  \bibinfo{volume}{9}, \bibinfo{number}{11} (\bibinfo{year}{2009}),
  \bibinfo{pages}{10--10}.
\newblock


\bibitem[\protect\citeauthoryear{Beck and Prazdny}{Beck and Prazdny}{1981}]%
        {beck1981highlights}
\bibfield{author}{\bibinfo{person}{Jacob Beck} {and} \bibinfo{person}{Slava
  Prazdny}.} \bibinfo{year}{1981}\natexlab{}.
\newblock \showarticletitle{Highlights and the perception of glossiness}.
\newblock \bibinfo{journal}{\emph{Attention, Perception, \& Psychophysics}}
  \bibinfo{volume}{30}, \bibinfo{number}{4} (\bibinfo{year}{1981}),
  \bibinfo{pages}{407--410}.
\newblock


\bibitem[\protect\citeauthoryear{Bell, Upchurch, Snavely, and Bala}{Bell
  et~al\mbox{.}}{2015}]%
        {bell2015material}
\bibfield{author}{\bibinfo{person}{Sean Bell}, \bibinfo{person}{Paul Upchurch},
  \bibinfo{person}{Noah Snavely}, {and} \bibinfo{person}{Kavita Bala}.}
  \bibinfo{year}{2015}\natexlab{}.
\newblock \showarticletitle{Material recognition in the wild with the materials
  in context database}. In \bibinfo{booktitle}{\emph{Proceedings of the IEEE
  Conference on Computer Vision and Pattern Recognition (CVPR)}}.
  \bibinfo{pages}{3479--3487}.
\newblock


\bibitem[\protect\citeauthoryear{Bloj, Kersten, and Hurlbert}{Bloj
  et~al\mbox{.}}{1999}]%
        {bloj1999perception}
\bibfield{author}{\bibinfo{person}{Marina~G Bloj}, \bibinfo{person}{Daniel
  Kersten}, {and} \bibinfo{person}{Anya~C Hurlbert}.}
  \bibinfo{year}{1999}\natexlab{}.
\newblock \showarticletitle{Perception of three-dimensional shape influences
  colour perception through mutual illumination}.
\newblock \bibinfo{journal}{\emph{Nature}} \bibinfo{volume}{402},
  \bibinfo{number}{6764} (\bibinfo{year}{1999}), \bibinfo{pages}{877}.
\newblock


\bibitem[\protect\citeauthoryear{Bousseau, Chapoulie, Ramamoorthi, and
  Agrawala}{Bousseau et~al\mbox{.}}{2011}]%
        {bousseau2011optimizing}
\bibfield{author}{\bibinfo{person}{Adrien Bousseau},
  \bibinfo{person}{Emmanuelle Chapoulie}, \bibinfo{person}{Ravi Ramamoorthi},
  {and} \bibinfo{person}{Maneesh Agrawala}.} \bibinfo{year}{2011}\natexlab{}.
\newblock \showarticletitle{Optimizing environment maps for material
  depiction}. In \bibinfo{booktitle}{\emph{Computer Graphics Forum}},
  Vol.~\bibinfo{volume}{30}. Wiley Online Library, \bibinfo{pages}{1171--1180}.
\newblock


\bibitem[\protect\citeauthoryear{Brady and Oliva}{Brady and Oliva}{2012}]%
        {brady2012spatial}
\bibfield{author}{\bibinfo{person}{Timothy Brady} {and} \bibinfo{person}{Aude
  Oliva}.} \bibinfo{year}{2012}\natexlab{}.
\newblock \showarticletitle{Spatial frequency integration during active
  perception: perceptual hysteresis when an object recedes}.
\newblock \bibinfo{journal}{\emph{Frontiers in Psychology}}
  \bibinfo{volume}{3} (\bibinfo{year}{2012}), \bibinfo{pages}{462}.
\newblock


\bibitem[\protect\citeauthoryear{Brossier, Bello, and Plumbley}{Brossier
  et~al\mbox{.}}{2004}]%
        {brossier2004real}
\bibfield{author}{\bibinfo{person}{Paul Brossier}, \bibinfo{person}{Juan~Pablo
  Bello}, {and} \bibinfo{person}{Mark~D Plumbley}.}
  \bibinfo{year}{2004}\natexlab{}.
\newblock \showarticletitle{Real-time temporal segmentation of note objects in
  music signals}. In \bibinfo{booktitle}{\emph{Proceedings of ICMC 2004, the
  30th Annual International Computer Music Conference}}.
\newblock


\bibitem[\protect\citeauthoryear{Burt and Adelson}{Burt and Adelson}{1983}]%
        {burt1983laplacian}
\bibfield{author}{\bibinfo{person}{Peter Burt} {and} \bibinfo{person}{Edward
  Adelson}.} \bibinfo{year}{1983}\natexlab{}.
\newblock \showarticletitle{The Laplacian pyramid as a compact image code}.
\newblock \bibinfo{journal}{\emph{IEEE Transactions on communications}}
  \bibinfo{volume}{31}, \bibinfo{number}{4} (\bibinfo{year}{1983}),
  \bibinfo{pages}{532--540}.
\newblock


\bibitem[\protect\citeauthoryear{Delanoy, Lagunas, Galve, Gutierrez, Serrano,
  Fleming, and Masia}{Delanoy et~al\mbox{.}}{2020}]%
        {delanoy2020role}
\bibfield{author}{\bibinfo{person}{Johanna Delanoy}, \bibinfo{person}{Manuel
  Lagunas}, \bibinfo{person}{Ignacio Galve}, \bibinfo{person}{Diego Gutierrez},
  \bibinfo{person}{Ana Serrano}, \bibinfo{person}{Roland Fleming}, {and}
  \bibinfo{person}{Belen Masia}.} \bibinfo{year}{2020}\natexlab{}.
\newblock \showarticletitle{The Role of Objective and Subjective Measures in
  Material Similarity Learning}. In \bibinfo{booktitle}{\emph{ACM SIGGRAPH 2020
  Posters}}. Article \bibinfo{articleno}{51}, \bibinfo{numpages}{2}~pages.
\newblock


\bibitem[\protect\citeauthoryear{Deng, Dong, Socher, Li, Li, and Fei-Fei}{Deng
  et~al\mbox{.}}{2009}]%
        {deng2009imagenet}
\bibfield{author}{\bibinfo{person}{Jia Deng}, \bibinfo{person}{Wei Dong},
  \bibinfo{person}{Richard Socher}, \bibinfo{person}{Li-Jia Li},
  \bibinfo{person}{Kai Li}, {and} \bibinfo{person}{Li Fei-Fei}.}
  \bibinfo{year}{2009}\natexlab{}.
\newblock \showarticletitle{Imagenet: A large-scale hierarchical image
  database}. In \bibinfo{booktitle}{\emph{Proceedings of the IEEE Conference on
  Computer Vision and Pattern Recognition (CVPR)}}. Ieee,
  \bibinfo{pages}{248--255}.
\newblock


\bibitem[\protect\citeauthoryear{Dror, Adelson, and Willsky}{Dror
  et~al\mbox{.}}{2001a}]%
        {dror2001estimating}
\bibfield{author}{\bibinfo{person}{Ron~O Dror}, \bibinfo{person}{Edward~H
  Adelson}, {and} \bibinfo{person}{Alan~S Willsky}.}
  \bibinfo{year}{2001}\natexlab{a}.
\newblock \showarticletitle{Estimating surface reflectance properties from
  images under unknown illumination}. In \bibinfo{booktitle}{\emph{Human Vision
  and Electronic Imaging VI}}, Vol.~\bibinfo{volume}{4299}. International
  Society for Optics and Photonics, \bibinfo{pages}{231--243}.
\newblock


\bibitem[\protect\citeauthoryear{Dror, Adelson, and Willsky}{Dror
  et~al\mbox{.}}{2001b}]%
        {dror2001surface}
\bibfield{author}{\bibinfo{person}{Ron~O Dror}, \bibinfo{person}{Edward~H
  Adelson}, {and} \bibinfo{person}{Alan~S Willsky}.}
  \bibinfo{year}{2001}\natexlab{b}.
\newblock \showarticletitle{Surface reflectance estimation and natural
  illumination statistics}.
\newblock


\bibitem[\protect\citeauthoryear{Faul}{Faul}{2019}]%
        {faul2019influence}
\bibfield{author}{\bibinfo{person}{Franz Faul}.}
  \bibinfo{year}{2019}\natexlab{}.
\newblock \showarticletitle{The influence of Fresnel effects on gloss
  perception}.
\newblock \bibinfo{journal}{\emph{Journal of Vision (JOV)}}
  \bibinfo{volume}{19}, \bibinfo{number}{13} (\bibinfo{year}{2019}),
  \bibinfo{pages}{1--1}.
\newblock


\bibitem[\protect\citeauthoryear{Filip, Chantler, Green, and Haindl}{Filip
  et~al\mbox{.}}{2008}]%
        {filip2008psychophysically}
\bibfield{author}{\bibinfo{person}{Jir{\'\i} Filip}, \bibinfo{person}{Mike~J
  Chantler}, \bibinfo{person}{Patrick~R Green}, {and} \bibinfo{person}{Michal
  Haindl}.} \bibinfo{year}{2008}\natexlab{}.
\newblock \showarticletitle{A psychophysically validated metric for
  bidirectional texture data reduction.}
\newblock \bibinfo{journal}{\emph{ACM Transactions on Graphics (TOG)}}
  \bibinfo{volume}{27}, \bibinfo{number}{5} (\bibinfo{year}{2008}),
  \bibinfo{pages}{138--1}.
\newblock


\bibitem[\protect\citeauthoryear{Fischer, {\v{S}}roubek, Perrinet, Redondo, and
  Crist{\'o}bal}{Fischer et~al\mbox{.}}{2007}]%
        {fischer2007self}
\bibfield{author}{\bibinfo{person}{Sylvain Fischer}, \bibinfo{person}{Filip
  {\v{S}}roubek}, \bibinfo{person}{Laurent Perrinet}, \bibinfo{person}{Rafael
  Redondo}, {and} \bibinfo{person}{Gabriel Crist{\'o}bal}.}
  \bibinfo{year}{2007}\natexlab{}.
\newblock \showarticletitle{Self-invertible 2D log-Gabor wavelets}.
\newblock \bibinfo{journal}{\emph{International Journal of Computer Vision}}
  \bibinfo{volume}{75}, \bibinfo{number}{2} (\bibinfo{year}{2007}),
  \bibinfo{pages}{231--246}.
\newblock


\bibitem[\protect\citeauthoryear{Fleming}{Fleming}{2014}]%
        {fleming2014}
\bibfield{author}{\bibinfo{person}{Roland~W. Fleming}.}
  \bibinfo{year}{2014}\natexlab{}.
\newblock \showarticletitle{Visual perception of materials and their
  properties}.
\newblock \bibinfo{journal}{\emph{Vision Research}}  \bibinfo{volume}{94}
  (\bibinfo{year}{2014}), \bibinfo{pages}{62 -- 75}.
\newblock


\bibitem[\protect\citeauthoryear{Fleming and B{\"u}lthoff}{Fleming and
  B{\"u}lthoff}{2005}]%
        {fleming2005low}
\bibfield{author}{\bibinfo{person}{Roland~W Fleming} {and}
  \bibinfo{person}{Heinrich~H B{\"u}lthoff}.} \bibinfo{year}{2005}\natexlab{}.
\newblock \showarticletitle{Low-level image cues in the perception of
  translucent materials}.
\newblock \bibinfo{journal}{\emph{ACM Transactions on Applied Perception
  (TAP)}} \bibinfo{volume}{2}, \bibinfo{number}{3} (\bibinfo{year}{2005}),
  \bibinfo{pages}{346--382}.
\newblock


\bibitem[\protect\citeauthoryear{Fleming, Dror, and Adelson}{Fleming
  et~al\mbox{.}}{2001}]%
        {fleming2001humans}
\bibfield{author}{\bibinfo{person}{Roland~W Fleming}, \bibinfo{person}{Ron~O
  Dror}, {and} \bibinfo{person}{Edward~H Adelson}.}
  \bibinfo{year}{2001}\natexlab{}.
\newblock \showarticletitle{How do humans determine reflectance properties
  under unknown illumination?}
\newblock  (\bibinfo{year}{2001}).
\newblock


\bibitem[\protect\citeauthoryear{Fleming, Dror, and Adelson}{Fleming
  et~al\mbox{.}}{2003}]%
        {fleming2003real}
\bibfield{author}{\bibinfo{person}{Roland~W Fleming}, \bibinfo{person}{Ron~O
  Dror}, {and} \bibinfo{person}{Edward~H Adelson}.}
  \bibinfo{year}{2003}\natexlab{}.
\newblock \showarticletitle{Real-world illumination and the perception of
  surface reflectance properties}.
\newblock \bibinfo{journal}{\emph{Journal of Vision (JOV)}}
  \bibinfo{volume}{3}, \bibinfo{number}{5} (\bibinfo{year}{2003}),
  \bibinfo{pages}{3--3}.
\newblock


\bibitem[\protect\citeauthoryear{Fleming, Gegenfurtner, and Nishida}{Fleming
  et~al\mbox{.}}{2015a}]%
        {fleming2015short}
\bibfield{author}{\bibinfo{person}{Roland~W Fleming}, \bibinfo{person}{Karl~R
  Gegenfurtner}, {and} \bibinfo{person}{Shin'ya Nishida}.}
  \bibinfo{year}{2015}\natexlab{a}.
\newblock \showarticletitle{Visual perception of materials: the science of
  stuff}.
\newblock \bibinfo{journal}{\emph{Vision Research}} \bibinfo{number}{109}
  (\bibinfo{year}{2015}), \bibinfo{pages}{123--124}.
\newblock


\bibitem[\protect\citeauthoryear{Fleming, Nishida, and Gegenfurtner}{Fleming
  et~al\mbox{.}}{2015b}]%
        {fleming2015}
\bibfield{author}{\bibinfo{person}{Roland~W. Fleming}, \bibinfo{person}{Shin'ya
  Nishida}, {and} \bibinfo{person}{Karl~R. Gegenfurtner}.}
  \bibinfo{year}{2015}\natexlab{b}.
\newblock \showarticletitle{Perception of material properties}.
\newblock \bibinfo{journal}{\emph{Vision Research}}  \bibinfo{volume}{115}
  (\bibinfo{year}{2015}), \bibinfo{pages}{157 -- 162}.
\newblock


\bibitem[\protect\citeauthoryear{Fleming and Storrs}{Fleming and
  Storrs}{2019}]%
        {fleming2019learning}
\bibfield{author}{\bibinfo{person}{Roland~W Fleming} {and}
  \bibinfo{person}{Katherine~R Storrs}.} \bibinfo{year}{2019}\natexlab{}.
\newblock \showarticletitle{Learning to see stuff}.
\newblock \bibinfo{journal}{\emph{Current Opinion in Behavioral Sciences}}
  \bibinfo{volume}{30} (\bibinfo{year}{2019}), \bibinfo{pages}{100--108}.
\newblock


\bibitem[\protect\citeauthoryear{Fleming, Wiebel, and Gegenfurtner}{Fleming
  et~al\mbox{.}}{2013}]%
        {fleming2013perceptual}
\bibfield{author}{\bibinfo{person}{Roland~W Fleming},
  \bibinfo{person}{Christiane Wiebel}, {and} \bibinfo{person}{Karl
  Gegenfurtner}.} \bibinfo{year}{2013}\natexlab{}.
\newblock \showarticletitle{Perceptual qualities and material classes}.
\newblock \bibinfo{journal}{\emph{Journal of Vision (JOV)}}
  \bibinfo{volume}{13}, \bibinfo{number}{8} (\bibinfo{year}{2013}),
  \bibinfo{pages}{9--9}.
\newblock


\bibitem[\protect\citeauthoryear{Garces, Agarwala, Gutierrez, and
  Hertzmann}{Garces et~al\mbox{.}}{2014}]%
        {GarcesSIG2014}
\bibfield{author}{\bibinfo{person}{Elena Garces}, \bibinfo{person}{Aseem
  Agarwala}, \bibinfo{person}{Diego Gutierrez}, {and} \bibinfo{person}{Aaron
  Hertzmann}.} \bibinfo{year}{2014}\natexlab{}.
\newblock \showarticletitle{A Similarity Measure for Illustration Style}.
\newblock \bibinfo{journal}{\emph{ACM Transactions on Graphics (Proc. SIGGRAPH
  2014)}} \bibinfo{volume}{33}, \bibinfo{number}{4} (\bibinfo{year}{2014}).
\newblock


\bibitem[\protect\citeauthoryear{Ged, Obein, Silvestri, Le~Rohellec, and
  Vi{\'e}not}{Ged et~al\mbox{.}}{2010}]%
        {ged2010recognizing}
\bibfield{author}{\bibinfo{person}{Guillaume Ged}, \bibinfo{person}{Ga{\"e}l
  Obein}, \bibinfo{person}{Zaccaria Silvestri}, \bibinfo{person}{Jean
  Le~Rohellec}, {and} \bibinfo{person}{Fran{\c{c}}oise Vi{\'e}not}.}
  \bibinfo{year}{2010}\natexlab{}.
\newblock \showarticletitle{Recognizing real materials from their glossy
  appearance}.
\newblock \bibinfo{journal}{\emph{Journal of Vision (JOV)}}
  \bibinfo{volume}{10}, \bibinfo{number}{9} (\bibinfo{year}{2010}),
  \bibinfo{pages}{18--18}.
\newblock


\bibitem[\protect\citeauthoryear{Geisler}{Geisler}{2008}]%
        {geisler2008visual}
\bibfield{author}{\bibinfo{person}{Wilson~S Geisler}.}
  \bibinfo{year}{2008}\natexlab{}.
\newblock \showarticletitle{Visual perception and the statistical properties of
  natural scenes}.
\newblock \bibinfo{journal}{\emph{Annual Review of Psychology}}
  \bibinfo{volume}{59} (\bibinfo{year}{2008}), \bibinfo{pages}{167--192}.
\newblock


\bibitem[\protect\citeauthoryear{Giesel and Zaidi}{Giesel and Zaidi}{2013}]%
        {giesel2013frequency}
\bibfield{author}{\bibinfo{person}{Martin Giesel} {and} \bibinfo{person}{Qasim
  Zaidi}.} \bibinfo{year}{2013}\natexlab{}.
\newblock \showarticletitle{Frequency-based heuristics for material
  perception}.
\newblock \bibinfo{journal}{\emph{Journal of Vision (JOV)}}
  \bibinfo{volume}{13}, \bibinfo{number}{14} (\bibinfo{year}{2013}),
  \bibinfo{pages}{7--7}.
\newblock


\bibitem[\protect\citeauthoryear{Guarnera, Guarnera, Toscani, Glencross, Li,
  Hardeberg, and Gegenfurtner}{Guarnera et~al\mbox{.}}{2018}]%
        {guarnera2018perceptually}
\bibfield{author}{\bibinfo{person}{Dar'ya Guarnera},
  \bibinfo{person}{Giuseppe~Claudio Guarnera}, \bibinfo{person}{Matteo
  Toscani}, \bibinfo{person}{Mashhuda Glencross}, \bibinfo{person}{Baihua Li},
  \bibinfo{person}{Jon~Yngve Hardeberg}, {and} \bibinfo{person}{Karl~R.
  Gegenfurtner}.} \bibinfo{year}{2018}\natexlab{}.
\newblock \showarticletitle{Perceptually Validated Analytical BRDFs Parameters
  Remapping}. In \bibinfo{booktitle}{\emph{ACM SIGGRAPH 2018 Talks}}. Article
  \bibinfo{articleno}{Article 17}, \bibinfo{numpages}{2}~pages.
\newblock


\bibitem[\protect\citeauthoryear{Guo, Guo, Pan, and Lu}{Guo
  et~al\mbox{.}}{2018}]%
        {guo2018brdf}
\bibfield{author}{\bibinfo{person}{Jie Guo}, \bibinfo{person}{Yanwen Guo},
  \bibinfo{person}{Jingui Pan}, {and} \bibinfo{person}{Wenzhou Lu}.}
  \bibinfo{year}{2018}\natexlab{}.
\newblock \showarticletitle{Brdf analysis with directional statistics and its
  applications}.
\newblock \bibinfo{journal}{\emph{IEEE Transactions on Visualization and
  Computer Graphics (TVCG)}} (\bibinfo{year}{2018}).
\newblock


\bibitem[\protect\citeauthoryear{Havran, Filip, and Myszkowski}{Havran
  et~al\mbox{.}}{2016}]%
        {havran2016}
\bibfield{author}{\bibinfo{person}{Vlastimil Havran}, \bibinfo{person}{Jiri
  Filip}, {and} \bibinfo{person}{Karol Myszkowski}.}
  \bibinfo{year}{2016}\natexlab{}.
\newblock \showarticletitle{{Perceptually Motivated {BRDF} Comparison using
  Single Image}}.
\newblock \bibinfo{journal}{\emph{Computer Graphics Forum}}
  (\bibinfo{year}{2016}).
\newblock
\showISSN{1467-8659}


\bibitem[\protect\citeauthoryear{Hawken and Parker}{Hawken and Parker}{1987}]%
        {hawken1987spatial}
\bibfield{author}{\bibinfo{person}{Michael~J Hawken} {and}
  \bibinfo{person}{Andrew~J Parker}.} \bibinfo{year}{1987}\natexlab{}.
\newblock \showarticletitle{Spatial properties of neurons in the monkey striate
  cortex}.
\newblock \bibinfo{journal}{\emph{Proceedings of the Royal society of London.
  Series B. Biological sciences}} \bibinfo{volume}{231}, \bibinfo{number}{1263}
  (\bibinfo{year}{1987}), \bibinfo{pages}{251--288}.
\newblock


\bibitem[\protect\citeauthoryear{He, Zhang, Ren, and Sun}{He
  et~al\mbox{.}}{2016}]%
        {he2016deep}
\bibfield{author}{\bibinfo{person}{Kaiming He}, \bibinfo{person}{Xiangyu
  Zhang}, \bibinfo{person}{Shaoqing Ren}, {and} \bibinfo{person}{Jian Sun}.}
  \bibinfo{year}{2016}\natexlab{}.
\newblock \showarticletitle{Deep residual learning for image recognition}. In
  \bibinfo{booktitle}{\emph{Proceedings of the IEEE Conference on Computer
  Vision and Pattern Recognition (CVPR)}}. \bibinfo{pages}{770--778}.
\newblock


\bibitem[\protect\citeauthoryear{Hunter et~al\mbox{.}}{Hunter
  et~al\mbox{.}}{1937}]%
        {hunter1937methods}
\bibfield{author}{\bibinfo{person}{Richard~S Hunter} {et~al\mbox{.}}}
  \bibinfo{year}{1937}\natexlab{}.
\newblock \showarticletitle{Methods of determining gloss}.
\newblock \bibinfo{journal}{\emph{NBS Research paper RP}}
  \bibinfo{volume}{958} (\bibinfo{year}{1937}).
\newblock


\bibitem[\protect\citeauthoryear{Jarabo, Wu, Dorsey, Rushmeier, and
  Gutierrez}{Jarabo et~al\mbox{.}}{2014}]%
        {jarabo2014btf}
\bibfield{author}{\bibinfo{person}{Adrian Jarabo}, \bibinfo{person}{Hongzhi
  Wu}, \bibinfo{person}{Julie Dorsey}, \bibinfo{person}{Holly Rushmeier}, {and}
  \bibinfo{person}{Diego Gutierrez}.} \bibinfo{year}{2014}\natexlab{}.
\newblock \showarticletitle{Effects of Approximate Filtering on the Appearance
  of Bidirectional Texture Functions}.
\newblock \bibinfo{journal}{\emph{{IEEE} Trans. on Visualization and Computer
  Graphics}} \bibinfo{volume}{20}, \bibinfo{number}{6} (\bibinfo{year}{2014}).
\newblock


\bibitem[\protect\citeauthoryear{Julesz}{Julesz}{1962}]%
        {julesz1962visual}
\bibfield{author}{\bibinfo{person}{Bela Julesz}.}
  \bibinfo{year}{1962}\natexlab{}.
\newblock \showarticletitle{Visual pattern discrimination}.
\newblock \bibinfo{journal}{\emph{IRE transactions on Information Theory}}
  \bibinfo{volume}{8}, \bibinfo{number}{2} (\bibinfo{year}{1962}),
  \bibinfo{pages}{84--92}.
\newblock


\bibitem[\protect\citeauthoryear{Kawato, Hayakawa, and Inui}{Kawato
  et~al\mbox{.}}{1993}]%
        {kawato1993forward}
\bibfield{author}{\bibinfo{person}{Mitsuo Kawato}, \bibinfo{person}{Hideki
  Hayakawa}, {and} \bibinfo{person}{Toshio Inui}.}
  \bibinfo{year}{1993}\natexlab{}.
\newblock \showarticletitle{A forward-inverse optics model of reciprocal
  connections between visual cortical areas}.
\newblock \bibinfo{journal}{\emph{Network: Computation in Neural Systems}}
  \bibinfo{volume}{4}, \bibinfo{number}{4} (\bibinfo{year}{1993}),
  \bibinfo{pages}{415--422}.
\newblock


\bibitem[\protect\citeauthoryear{Kerr and Pellacini}{Kerr and
  Pellacini}{2010}]%
        {kerr2010toward}
\bibfield{author}{\bibinfo{person}{William~B Kerr} {and} \bibinfo{person}{Fabio
  Pellacini}.} \bibinfo{year}{2010}\natexlab{}.
\newblock \showarticletitle{Toward evaluating material design interface
  paradigms for novice users}. In \bibinfo{booktitle}{\emph{ACM Transactions on
  Graphics (TOG)}}, Vol.~\bibinfo{volume}{29}. ACM, \bibinfo{pages}{35}.
\newblock


\bibitem[\protect\citeauthoryear{Kersten, Mamassian, and Yuille}{Kersten
  et~al\mbox{.}}{2004}]%
        {kersten2004object}
\bibfield{author}{\bibinfo{person}{Daniel Kersten}, \bibinfo{person}{Pascal
  Mamassian}, {and} \bibinfo{person}{Alan Yuille}.}
  \bibinfo{year}{2004}\natexlab{}.
\newblock \showarticletitle{Object perception as Bayesian inference}.
\newblock \bibinfo{journal}{\emph{Annual Review of Psychology}}
  \bibinfo{volume}{55} (\bibinfo{year}{2004}), \bibinfo{pages}{271--304}.
\newblock


\bibitem[\protect\citeauthoryear{Kim and Anderson}{Kim and Anderson}{2010}]%
        {kim2010image}
\bibfield{author}{\bibinfo{person}{Juno Kim} {and} \bibinfo{person}{Barton~L
  Anderson}.} \bibinfo{year}{2010}\natexlab{}.
\newblock \showarticletitle{Image statistics and the perception of surface
  gloss and lightness}.
\newblock \bibinfo{journal}{\emph{Journal of Vision (JOV)}}
  \bibinfo{volume}{10}, \bibinfo{number}{9} (\bibinfo{year}{2010}),
  \bibinfo{pages}{3--3}.
\newblock


\bibitem[\protect\citeauthoryear{Kingma and Ba}{Kingma and Ba}{2014}]%
        {kingma2014}
\bibfield{author}{\bibinfo{person}{Diederik~P Kingma} {and}
  \bibinfo{person}{Jimmy Ba}.} \bibinfo{year}{2014}\natexlab{}.
\newblock \showarticletitle{Adam: A method for stochastic optimization}.
\newblock \bibinfo{journal}{\emph{arXiv preprint arXiv:1412.6980}}
  (\bibinfo{year}{2014}).
\newblock


\bibitem[\protect\citeauthoryear{Lagunas, Garces, and Gutierrez}{Lagunas
  et~al\mbox{.}}{2018}]%
        {lagunas2018}
\bibfield{author}{\bibinfo{person}{Manuel Lagunas}, \bibinfo{person}{Elena
  Garces}, {and} \bibinfo{person}{Diego Gutierrez}.}
  \bibinfo{year}{2018}\natexlab{}.
\newblock \showarticletitle{Learning icons appearance similarity}.
\newblock \bibinfo{journal}{\emph{Multimedia Tools and Applications}}
  (\bibinfo{year}{2018}), \bibinfo{pages}{1--19}.
\newblock


\bibitem[\protect\citeauthoryear{Lagunas, Malpica, Serrano, Garces, Gutierrez,
  and Masia}{Lagunas et~al\mbox{.}}{2019}]%
        {lagunas2019similarity}
\bibfield{author}{\bibinfo{person}{Manuel Lagunas}, \bibinfo{person}{Sandra
  Malpica}, \bibinfo{person}{Ana Serrano}, \bibinfo{person}{Elena Garces},
  \bibinfo{person}{Diego Gutierrez}, {and} \bibinfo{person}{Belen Masia}.}
  \bibinfo{year}{2019}\natexlab{}.
\newblock \showarticletitle{A Similarity Measure for Material Appearance}.
\newblock \bibinfo{journal}{\emph{ACM Transactions on Graphics (Proc. SIGGRAPH
  2019)}} \bibinfo{volume}{38}, \bibinfo{number}{4} (\bibinfo{year}{2019}).
\newblock


\bibitem[\protect\citeauthoryear{Lee and Lee}{Lee and Lee}{2016}]%
        {lee2016fusion}
\bibfield{author}{\bibinfo{person}{Seohyung Lee} {and} \bibinfo{person}{Daeho
  Lee}.} \bibinfo{year}{2016}\natexlab{}.
\newblock \showarticletitle{Fusion of IR and Visual Images Based on Gaussian
  and Laplacian Decomposition Using Histogram Distributions and Edge
  Selection}.
\newblock \bibinfo{journal}{\emph{Mathematical Problems in Engineering}}
  \bibinfo{volume}{2016} (\bibinfo{year}{2016}).
\newblock


\bibitem[\protect\citeauthoryear{Leloup, Pointer, Dutr{\'e}, and
  Hanselaer}{Leloup et~al\mbox{.}}{2010}]%
        {leloup2010geometry}
\bibfield{author}{\bibinfo{person}{Fr{\'e}d{\'e}ric~B Leloup},
  \bibinfo{person}{Michael~R Pointer}, \bibinfo{person}{Philip Dutr{\'e}},
  {and} \bibinfo{person}{Peter Hanselaer}.} \bibinfo{year}{2010}\natexlab{}.
\newblock \showarticletitle{Geometry of illumination, luminance contrast, and
  gloss perception}.
\newblock \bibinfo{journal}{\emph{JOSA A}} \bibinfo{volume}{27},
  \bibinfo{number}{9} (\bibinfo{year}{2010}), \bibinfo{pages}{2046--2054}.
\newblock


\bibitem[\protect\citeauthoryear{Li and Fritz}{Li and Fritz}{2012}]%
        {li2012recognizing}
\bibfield{author}{\bibinfo{person}{Wenbin Li} {and} \bibinfo{person}{Mario
  Fritz}.} \bibinfo{year}{2012}\natexlab{}.
\newblock \showarticletitle{Recognizing materials from virtual examples}. In
  \bibinfo{booktitle}{\emph{European Conference on Computer Vision (ECCV)}}.
  Springer, \bibinfo{pages}{345--358}.
\newblock


\bibitem[\protect\citeauthoryear{Maloney and Brainard}{Maloney and
  Brainard}{2010}]%
        {maloney2010}
\bibfield{author}{\bibinfo{person}{Laurence~T Maloney} {and}
  \bibinfo{person}{David~H Brainard}.} \bibinfo{year}{2010}\natexlab{}.
\newblock \showarticletitle{Color and material perception: Achievements and
  challenges}.
\newblock \bibinfo{journal}{\emph{Journal of Vision (JOV)}}
  \bibinfo{volume}{10}, \bibinfo{number}{9} (\bibinfo{year}{2010}),
  \bibinfo{pages}{19--19}.
\newblock


\bibitem[\protect\citeauthoryear{Mao, Lagunas, Masia, and Gutierrez}{Mao
  et~al\mbox{.}}{2019}]%
        {mao2019motion}
\bibfield{author}{\bibinfo{person}{Ruiquan Mao}, \bibinfo{person}{Manuel
  Lagunas}, \bibinfo{person}{Belen Masia}, {and} \bibinfo{person}{Diego
  Gutierrez}.} \bibinfo{year}{2019}\natexlab{}.
\newblock \showarticletitle{The Effect of Motion on the Perception of Material
  Appearance}. In \bibinfo{booktitle}{\emph{Proceedings of the ACM Symposium on
  Applied Perception (SAP)}}. \bibinfo{publisher}{ACM}, Article
  \bibinfo{articleno}{16}, \bibinfo{numpages}{9}~pages.
\newblock


\bibitem[\protect\citeauthoryear{Marlow, Kim, and Anderson}{Marlow
  et~al\mbox{.}}{2012}]%
        {marlow2012perception}
\bibfield{author}{\bibinfo{person}{Phillip~J Marlow}, \bibinfo{person}{Juno
  Kim}, {and} \bibinfo{person}{Barton~L Anderson}.}
  \bibinfo{year}{2012}\natexlab{}.
\newblock \showarticletitle{The perception and misperception of specular
  surface reflectance}.
\newblock \bibinfo{journal}{\emph{Current Biology}} \bibinfo{volume}{22},
  \bibinfo{number}{20} (\bibinfo{year}{2012}), \bibinfo{pages}{1909--1913}.
\newblock


\bibitem[\protect\citeauthoryear{Matusik, Pfister, Brand, and McMillan}{Matusik
  et~al\mbox{.}}{2003}]%
        {matusik2003}
\bibfield{author}{\bibinfo{person}{Wojciech Matusik},
  \bibinfo{person}{Hanspeter Pfister}, \bibinfo{person}{Matt Brand}, {and}
  \bibinfo{person}{Leonard McMillan}.} \bibinfo{year}{2003}\natexlab{}.
\newblock \showarticletitle{A Data-Driven Reflectance Model}.
\newblock \bibinfo{journal}{\emph{ACM Transactions on Graphics (TOG)}}
  \bibinfo{volume}{22}, \bibinfo{number}{3} (\bibinfo{date}{July}
  \bibinfo{year}{2003}), \bibinfo{pages}{759--769}.
\newblock


\bibitem[\protect\citeauthoryear{McInnes and Healy}{McInnes and Healy}{2018}]%
        {mcinnes2018umap}
\bibfield{author}{\bibinfo{person}{Leland McInnes} {and} \bibinfo{person}{John
  Healy}.} \bibinfo{year}{2018}\natexlab{}.
\newblock \showarticletitle{Umap: Uniform manifold approximation and projection
  for dimension reduction}.
\newblock \bibinfo{journal}{\emph{arXiv preprint arXiv:1802.03426}}
  (\bibinfo{year}{2018}).
\newblock


\bibitem[\protect\citeauthoryear{Motoyoshi, Nishida, Sharan, and
  Adelson}{Motoyoshi et~al\mbox{.}}{2007}]%
        {motoyoshi2007image}
\bibfield{author}{\bibinfo{person}{Isamu Motoyoshi}, \bibinfo{person}{Shin'ya
  Nishida}, \bibinfo{person}{Lavanya Sharan}, {and} \bibinfo{person}{Edward~H
  Adelson}.} \bibinfo{year}{2007}\natexlab{}.
\newblock \showarticletitle{Image statistics and the perception of surface
  qualities}.
\newblock \bibinfo{journal}{\emph{Nature}} \bibinfo{volume}{447},
  \bibinfo{number}{7141} (\bibinfo{year}{2007}), \bibinfo{pages}{206--209}.
\newblock


\bibitem[\protect\citeauthoryear{Nagai, Matsushima, Koida, Tani, Kitazaki, and
  Nakauchi}{Nagai et~al\mbox{.}}{2015}]%
        {nagai2015}
\bibfield{author}{\bibinfo{person}{Takehiro Nagai}, \bibinfo{person}{Toshiki
  Matsushima}, \bibinfo{person}{Kowa Koida}, \bibinfo{person}{Yusuke Tani},
  \bibinfo{person}{Michiteru Kitazaki}, {and} \bibinfo{person}{Shigeki
  Nakauchi}.} \bibinfo{year}{2015}\natexlab{}.
\newblock \showarticletitle{Temporal properties of material categorization and
  material rating: visual vs non-visual material features}.
\newblock \bibinfo{journal}{\emph{Vision Research}}  \bibinfo{volume}{115}
  (\bibinfo{year}{2015}), \bibinfo{pages}{259 -- 270}.
\newblock
\showISSN{0042-6989}
\newblock
\shownote{Perception of Material Properties (Part II).}


\bibitem[\protect\citeauthoryear{Nishida and Shinya}{Nishida and
  Shinya}{1998}]%
        {nishida1998use}
\bibfield{author}{\bibinfo{person}{Shin’ya Nishida} {and}
  \bibinfo{person}{Mikio Shinya}.} \bibinfo{year}{1998}\natexlab{}.
\newblock \showarticletitle{Use of image-based information in judgments of
  surface-reflectance properties}.
\newblock \bibinfo{journal}{\emph{JOSA A}} \bibinfo{volume}{15},
  \bibinfo{number}{12} (\bibinfo{year}{1998}), \bibinfo{pages}{2951--2965}.
\newblock


\bibitem[\protect\citeauthoryear{Obein, Knoblauch, and Vi{\'e}ot}{Obein
  et~al\mbox{.}}{2004}]%
        {obein2004difference}
\bibfield{author}{\bibinfo{person}{Ga{\"e} Obein}, \bibinfo{person}{Kenneth
  Knoblauch}, {and} \bibinfo{person}{Fran{\c{c}}ise Vi{\'e}ot}.}
  \bibinfo{year}{2004}\natexlab{}.
\newblock \showarticletitle{Difference scaling of gloss: Nonlinearity,
  binocularity, and constancy}.
\newblock \bibinfo{journal}{\emph{Journal of Vision (JOV)}}
  \bibinfo{volume}{4}, \bibinfo{number}{9} (\bibinfo{year}{2004}),
  \bibinfo{pages}{4--4}.
\newblock


\bibitem[\protect\citeauthoryear{Oliva and Torralba}{Oliva and
  Torralba}{2001}]%
        {oliva2001modeling}
\bibfield{author}{\bibinfo{person}{Aude Oliva} {and} \bibinfo{person}{Antonio
  Torralba}.} \bibinfo{year}{2001}\natexlab{}.
\newblock \showarticletitle{Modeling the shape of the scene: A holistic
  representation of the spatial envelope}.
\newblock \bibinfo{journal}{\emph{International Journal of Computer Vision
  (IJCV)}} \bibinfo{volume}{42}, \bibinfo{number}{3} (\bibinfo{year}{2001}),
  \bibinfo{pages}{145--175}.
\newblock


\bibitem[\protect\citeauthoryear{Olkkonen and Brainard}{Olkkonen and
  Brainard}{2010}]%
        {olkkonen2010perceived}
\bibfield{author}{\bibinfo{person}{Maria Olkkonen} {and}
  \bibinfo{person}{David~H Brainard}.} \bibinfo{year}{2010}\natexlab{}.
\newblock \showarticletitle{Perceived glossiness and lightness under real-world
  illumination}.
\newblock \bibinfo{journal}{\emph{Journal of Vision (JOV)}}
  \bibinfo{volume}{10}, \bibinfo{number}{9} (\bibinfo{year}{2010}),
  \bibinfo{pages}{5--5}.
\newblock


\bibitem[\protect\citeauthoryear{Olkkonen and Brainard}{Olkkonen and
  Brainard}{2011}]%
        {olkkonen2011joint}
\bibfield{author}{\bibinfo{person}{Maria Olkkonen} {and}
  \bibinfo{person}{David~H Brainard}.} \bibinfo{year}{2011}\natexlab{}.
\newblock \showarticletitle{Joint effects of illumination geometry and object
  shape in the perception of surface reflectance}.
\newblock \bibinfo{journal}{\emph{i-Perception}} \bibinfo{volume}{2},
  \bibinfo{number}{9} (\bibinfo{year}{2011}), \bibinfo{pages}{1014--1034}.
\newblock


\bibitem[\protect\citeauthoryear{Palmer}{Palmer}{1975}]%
        {palmer1975}
\bibfield{author}{\bibinfo{person}{Stephen~E Palmer}.}
  \bibinfo{year}{1975}\natexlab{}.
\newblock \showarticletitle{Visual perception and world knowledge: Notes on a
  model of sensory-cognitive interaction}.
\newblock \bibinfo{journal}{\emph{Explorations in Cognition}}
  (\bibinfo{year}{1975}), \bibinfo{pages}{279--307}.
\newblock


\bibitem[\protect\citeauthoryear{Pele and Werman}{Pele and Werman}{2010}]%
        {pele2010quadratic}
\bibfield{author}{\bibinfo{person}{Ofir Pele} {and} \bibinfo{person}{Michael
  Werman}.} \bibinfo{year}{2010}\natexlab{}.
\newblock \showarticletitle{The quadratic-chi histogram distance family}. In
  \bibinfo{booktitle}{\emph{European conference on computer vision}}. Springer,
  \bibinfo{pages}{749--762}.
\newblock


\bibitem[\protect\citeauthoryear{Pizlo}{Pizlo}{2001}]%
        {pizlo2001perception}
\bibfield{author}{\bibinfo{person}{Zygmunt Pizlo}.}
  \bibinfo{year}{2001}\natexlab{}.
\newblock \showarticletitle{Perception viewed as an inverse problem}.
\newblock \bibinfo{journal}{\emph{Vision Research}} \bibinfo{volume}{41},
  \bibinfo{number}{24} (\bibinfo{year}{2001}), \bibinfo{pages}{3145--3161}.
\newblock


\bibitem[\protect\citeauthoryear{Ramamoorthi and Hanrahan}{Ramamoorthi and
  Hanrahan}{2001}]%
        {ramamoorthi2001efficient}
\bibfield{author}{\bibinfo{person}{Ravi Ramamoorthi} {and} \bibinfo{person}{Pat
  Hanrahan}.} \bibinfo{year}{2001}\natexlab{}.
\newblock \showarticletitle{An efficient representation for irradiance
  environment maps}. In \bibinfo{booktitle}{\emph{Proceedings of the Annual
  conference on Computer Graphics and Interactive Techniques}}.
  \bibinfo{pages}{497--500}.
\newblock


\bibitem[\protect\citeauthoryear{Rubinstein, Gutierrez, Sorkine, and
  Shamir}{Rubinstein et~al\mbox{.}}{2010}]%
        {Rubinstein10Comparative}
\bibfield{author}{\bibinfo{person}{Michael Rubinstein}, \bibinfo{person}{Diego
  Gutierrez}, \bibinfo{person}{Olga Sorkine}, {and} \bibinfo{person}{Ariel
  Shamir}.} \bibinfo{year}{2010}\natexlab{}.
\newblock \showarticletitle{A Comparative Study of Image Retargeting}.
\newblock \bibinfo{journal}{\emph{ACM Transactions on Graphics (Proc. SIGGRAPH
  Asia 2010)}} \bibinfo{volume}{29}, \bibinfo{number}{6}
  (\bibinfo{year}{2010}), \bibinfo{pages}{160:1--160:10}.
\newblock


\bibitem[\protect\citeauthoryear{Schaffalitzky and Zisserman}{Schaffalitzky and
  Zisserman}{2001}]%
        {schaffalitzky2001viewpoint}
\bibfield{author}{\bibinfo{person}{Frederik Schaffalitzky} {and}
  \bibinfo{person}{Andrew Zisserman}.} \bibinfo{year}{2001}\natexlab{}.
\newblock \showarticletitle{Viewpoint invariant texture matching and wide
  baseline stereo}. In \bibinfo{booktitle}{\emph{Proceedings of the IEEE
  International Conference on Computer Vision (ICCV)}},
  Vol.~\bibinfo{volume}{2}. IEEE, \bibinfo{pages}{636--643}.
\newblock


\bibitem[\protect\citeauthoryear{Schiller, Finlay, and Volman}{Schiller
  et~al\mbox{.}}{1976}]%
        {schiller1976quantitative}
\bibfield{author}{\bibinfo{person}{Peter~H Schiller},
  \bibinfo{person}{Barbara~L Finlay}, {and} \bibinfo{person}{Susan~F Volman}.}
  \bibinfo{year}{1976}\natexlab{}.
\newblock \showarticletitle{Quantitative studies of single-cell properties in
  monkey striate cortex. I. Spatiotemporal organization of receptive fields}.
\newblock \bibinfo{journal}{\emph{Journal of neurophysiology}}
  \bibinfo{volume}{39}, \bibinfo{number}{6} (\bibinfo{year}{1976}),
  \bibinfo{pages}{1288--1319}.
\newblock


\bibitem[\protect\citeauthoryear{Schl{\"u}ter and Faul}{Schl{\"u}ter and
  Faul}{2019}]%
        {schluter2019visual}
\bibfield{author}{\bibinfo{person}{Nick Schl{\"u}ter} {and}
  \bibinfo{person}{Franz Faul}.} \bibinfo{year}{2019}\natexlab{}.
\newblock \showarticletitle{Visual shape perception in the case of transparent
  objects}.
\newblock \bibinfo{journal}{\emph{Journal of Vision (JOV)}}
  \bibinfo{volume}{19}, \bibinfo{number}{4} (\bibinfo{year}{2019}),
  \bibinfo{pages}{24--24}.
\newblock


\bibitem[\protect\citeauthoryear{Serrano, Gutierrez, Myszkowski, Seidel, and
  Masia}{Serrano et~al\mbox{.}}{2016}]%
        {serrano2016}
\bibfield{author}{\bibinfo{person}{Ana Serrano}, \bibinfo{person}{Diego
  Gutierrez}, \bibinfo{person}{Karol Myszkowski}, \bibinfo{person}{Hans-Peter
  Seidel}, {and} \bibinfo{person}{Belen Masia}.}
  \bibinfo{year}{2016}\natexlab{}.
\newblock \showarticletitle{An Intuitive Control Space for Material
  Appearance}.
\newblock \bibinfo{journal}{\emph{ACM Transactions on Graphics (TOG)}}
  \bibinfo{volume}{35}, \bibinfo{number}{6}, Article \bibinfo{articleno}{186}
  (\bibinfo{date}{Nov.} \bibinfo{year}{2016}),
  \bibinfo{numpages}{186:1--186:12}~pages.
\newblock
\showISSN{0730-0301}


\bibitem[\protect\citeauthoryear{S{\`e}ve}{S{\`e}ve}{1993}]%
        {seve1993problems}
\bibfield{author}{\bibinfo{person}{Robert S{\`e}ve}.}
  \bibinfo{year}{1993}\natexlab{}.
\newblock \showarticletitle{Problems connected with the concept of gloss}.
\newblock \bibinfo{journal}{\emph{Color Research \& Application}}
  \bibinfo{volume}{18}, \bibinfo{number}{4} (\bibinfo{year}{1993}),
  \bibinfo{pages}{241--252}.
\newblock


\bibitem[\protect\citeauthoryear{Sharan, Rosenholtz, and Adelson}{Sharan
  et~al\mbox{.}}{2009}]%
        {sharan2009}
\bibfield{author}{\bibinfo{person}{Lavanya Sharan}, \bibinfo{person}{Ruth
  Rosenholtz}, {and} \bibinfo{person}{Edward Adelson}.}
  \bibinfo{year}{2009}\natexlab{}.
\newblock \showarticletitle{Material perception: What can you see in a brief
  glance?}
\newblock \bibinfo{journal}{\emph{Journal of Vision (JOV)}}
  \bibinfo{volume}{9}, \bibinfo{number}{8} (\bibinfo{year}{2009}),
  \bibinfo{pages}{784--784}.
\newblock


\bibitem[\protect\citeauthoryear{Sharan, Rosenholtz, and Adelson}{Sharan
  et~al\mbox{.}}{2008}]%
        {sharan2008eye}
\bibfield{author}{\bibinfo{person}{Lavanya Sharan}, \bibinfo{person}{Ruth
  Rosenholtz}, {and} \bibinfo{person}{Edward~H Adelson}.}
  \bibinfo{year}{2008}\natexlab{}.
\newblock \showarticletitle{Eye movements for shape and material perception}.
\newblock \bibinfo{journal}{\emph{Journal of Vision (JOV)}}
  \bibinfo{volume}{8}, \bibinfo{number}{6} (\bibinfo{year}{2008}),
  \bibinfo{pages}{219--219}.
\newblock


\bibitem[\protect\citeauthoryear{Sun, Serrano, Gutierrez, and Masia}{Sun
  et~al\mbox{.}}{2017}]%
        {sun2017attribute}
\bibfield{author}{\bibinfo{person}{Tiancheng Sun}, \bibinfo{person}{Ana
  Serrano}, \bibinfo{person}{Diego Gutierrez}, {and} \bibinfo{person}{Belen
  Masia}.} \bibinfo{year}{2017}\natexlab{}.
\newblock \showarticletitle{Attribute-preserving gamut mapping of measured
  BRDFs}. In \bibinfo{booktitle}{\emph{Computer Graphics Forum}},
  Vol.~\bibinfo{volume}{36}. Wiley Online Library, \bibinfo{pages}{47--54}.
\newblock


\bibitem[\protect\citeauthoryear{Szegedy, Vanhoucke, Ioffe, Shlens, and
  Wojna}{Szegedy et~al\mbox{.}}{2015}]%
        {szegedy2015}
\bibfield{author}{\bibinfo{person}{Christian Szegedy}, \bibinfo{person}{Vincent
  Vanhoucke}, \bibinfo{person}{Sergey Ioffe}, \bibinfo{person}{Jonathon
  Shlens}, {and} \bibinfo{person}{Zbigniew Wojna}.}
  \bibinfo{year}{2015}\natexlab{}.
\newblock \showarticletitle{Rethinking the inception architecture for computer
  vision. arXiv}.
\newblock  (\bibinfo{year}{2015}).
\newblock


\bibitem[\protect\citeauthoryear{Thompson, Fleming, Creem-Regehr, and
  Stefanucci}{Thompson et~al\mbox{.}}{2011}]%
        {Thompson2011}
\bibfield{author}{\bibinfo{person}{William Thompson}, \bibinfo{person}{Roland
  Fleming}, \bibinfo{person}{Sarah Creem-Regehr}, {and}
  \bibinfo{person}{Jeanine~Kelly Stefanucci}.} \bibinfo{year}{2011}\natexlab{}.
\newblock \bibinfo{booktitle}{\emph{Visual Perception from a Computer Graphics
  Perspective} (\bibinfo{edition}{1st} ed.)}.
\newblock \bibinfo{publisher}{A. K. Peters, Ltd.}, \bibinfo{address}{Natick,
  MA, USA}.
\newblock
\showISBNx{1568814658, 9781568814650}


\bibitem[\protect\citeauthoryear{Thompson, Fleming, Creem-Regehr, and
  Stefanucci}{Thompson et~al\mbox{.}}{2016}]%
        {thompson2016visual}
\bibfield{author}{\bibinfo{person}{William Thompson}, \bibinfo{person}{Roland
  Fleming}, \bibinfo{person}{Sarah Creem-Regehr}, {and}
  \bibinfo{person}{Jeanine~Kelly Stefanucci}.} \bibinfo{year}{2016}\natexlab{}.
\newblock \bibinfo{booktitle}{\emph{Visual perception from a computer graphics
  perspective}}.
\newblock \bibinfo{publisher}{AK Peters/CRC Press}.
\newblock


\bibitem[\protect\citeauthoryear{Van Der~Maaten and Weinberger}{Van Der~Maaten
  and Weinberger}{2012}]%
        {van2012stochastic}
\bibfield{author}{\bibinfo{person}{Laurens Van Der~Maaten} {and}
  \bibinfo{person}{Kilian Weinberger}.} \bibinfo{year}{2012}\natexlab{}.
\newblock \showarticletitle{Stochastic triplet embedding}. In
  \bibinfo{booktitle}{\emph{Machine Learning for Signal Processing (MLSP), 2012
  IEEE International Workshop on}}. IEEE, \bibinfo{pages}{1--6}.
\newblock


\bibitem[\protect\citeauthoryear{Vangorp, Laurijssen, and Dutr{\'e}}{Vangorp
  et~al\mbox{.}}{2007}]%
        {vangorp2007}
\bibfield{author}{\bibinfo{person}{Peter Vangorp}, \bibinfo{person}{Jurgen
  Laurijssen}, {and} \bibinfo{person}{Philip Dutr{\'e}}.}
  \bibinfo{year}{2007}\natexlab{}.
\newblock \showarticletitle{The Influence of Shape on the Perception of
  Material Reflectance}.
\newblock \bibinfo{journal}{\emph{ACM Transactions on Graphics (TOG)}}
  \bibinfo{volume}{26}, \bibinfo{number}{3}, Article \bibinfo{articleno}{77}
  (\bibinfo{date}{July} \bibinfo{year}{2007}).
\newblock
\showISSN{0730-0301}


\bibitem[\protect\citeauthoryear{V{\'a}vra and Filip}{V{\'a}vra and
  Filip}{2016}]%
        {vavra2016minimal}
\bibfield{author}{\bibinfo{person}{Radom{\'\i}r V{\'a}vra} {and}
  \bibinfo{person}{Jir{\'\i} Filip}.} \bibinfo{year}{2016}\natexlab{}.
\newblock \showarticletitle{Minimal sampling for effective acquisition of
  anisotropic BRDFs}. In \bibinfo{booktitle}{\emph{Computer Graphics Forum}},
  Vol.~\bibinfo{volume}{35}. Wiley Online Library, \bibinfo{pages}{299--309}.
\newblock


\bibitem[\protect\citeauthoryear{Wang, Zhu, Hiroaki, Chandraker, Efros, and
  Ramamoorthi}{Wang et~al\mbox{.}}{2016}]%
        {wang20164d}
\bibfield{author}{\bibinfo{person}{Ting-Chun Wang}, \bibinfo{person}{Jun-Yan
  Zhu}, \bibinfo{person}{Ebi Hiroaki}, \bibinfo{person}{Manmohan Chandraker},
  \bibinfo{person}{Alexei~A Efros}, {and} \bibinfo{person}{Ravi Ramamoorthi}.}
  \bibinfo{year}{2016}\natexlab{}.
\newblock \showarticletitle{A 4D light-field dataset and CNN architectures for
  material recognition}. In \bibinfo{booktitle}{\emph{European Conference on
  Computer Vision}}. Springer, \bibinfo{pages}{121--138}.
\newblock


\bibitem[\protect\citeauthoryear{Welinder, Branson, Perona, and
  Belongie}{Welinder et~al\mbox{.}}{2010}]%
        {welinder2010multidimensional}
\bibfield{author}{\bibinfo{person}{Peter Welinder}, \bibinfo{person}{Steve
  Branson}, \bibinfo{person}{Pietro Perona}, {and} \bibinfo{person}{Serge~J
  Belongie}.} \bibinfo{year}{2010}\natexlab{}.
\newblock \showarticletitle{The multidimensional wisdom of crowds}. In
  \bibinfo{booktitle}{\emph{Advances in Neural Information Processing Systems
  (NeurIPS)}}. \bibinfo{pages}{2424--2432}.
\newblock


\bibitem[\protect\citeauthoryear{Zhang, de~Ridder, and Pont}{Zhang
  et~al\mbox{.}}{2015}]%
        {zhang2015influence}
\bibfield{author}{\bibinfo{person}{Fan Zhang}, \bibinfo{person}{Huib de
  Ridder}, {and} \bibinfo{person}{Sylvia Pont}.}
  \bibinfo{year}{2015}\natexlab{}.
\newblock \showarticletitle{The influence of lighting on visual perception of
  material qualities}. In \bibinfo{booktitle}{\emph{Human Vision and Electronic
  Imaging XX}}, Vol.~\bibinfo{volume}{9394}. International Society for Optics
  and Photonics, \bibinfo{pages}{93940Q}.
\newblock


\end{thebibliography}

\end{document}